\PassOptionsToPackage{table}{xcolor}
\documentclass{style}
\usepackage{graphicx}
\usepackage{amsmath}
\usepackage{booktabs}
\usepackage{multirow}
\usepackage{enumitem}
\usepackage{amssymb}
\usepackage{wrapfig}
\usepackage{float}
\usepackage{microtype}
\usepackage{tabularx}
\usepackage{pifont}
\usepackage[scaled=0.92]{zi4}
\setlist{leftmargin=*,itemsep=1pt,topsep=2pt,parsep=0pt}

\usepackage{xcolor}

\title{PyRUA-REAL: Robot Exploration and Adaptive Learning through RSI in the Real World}

\author{
  Qiang Wang\texorpdfstring{$^{*1}$}{} \hspace{0.1cm}
  Jianxin Bi\texorpdfstring{$^{*1,3\ddagger}$}{} \hspace{0.1cm}
  Shulong Jiang\texorpdfstring{$^{*1}$}{} \hspace{0.1cm}
  Ruiyang Si\texorpdfstring{$^{1}$}{} \hspace{0.1cm}
  Wenbo Huang\texorpdfstring{$^{1}$}{} \hspace{0.1cm}
  Kaiqi Chen\texorpdfstring{$^{1}$}{} \hspace{0.1cm}
  Jiaxin Li\texorpdfstring{$^{1}$}{} \hspace{0.1cm}
  Liyun Li\texorpdfstring{$^{2}$}{} \hspace{0.1cm}
  Xuanyi Li\texorpdfstring{$^{2\dagger}$}{}\hspace{0.1cm}
  Daquan Zhou\texorpdfstring{$^{1\dagger}$}{}
  
}

\affiliation[1]{Peking University}
\affiliation[2]{EngineAI}
\affiliation[3]{National University of Singapore}
\contribution[*]{Equal Contribution}
\contribution[\ddagger]{Project Lead}
\contribution[\dagger]{Corresponding author}

\abstract{
Coding agents powered by models such as GPT-6-Astra connect spatial reasoning with executable robot behavior, raising the question of how such agents can produce lasting improvements in robot capabilities from physical interaction.
In this work, we study embodied recursive self-improvement (RSI) through
\emph{real-world} exploration, code refinement, and deployment using PyRUA-REAL,
a framework in which coding agents retain interaction experience as
persistent memory and reusable executable skills that inform subsequent
exploration and improvement.
Our empirical study places physical manipulation at the center of the improvement process, examining how agents collaborate, what kinds of execution feedback support effective code refinement, and how task specifications and previously acquired skills influence robustness and task transfer.
We further investigate the efficiency and capability of the resulting programs when deployed without involving an online agent for decision-making.
\textbf{Across eight real-world tasks, PyRUA-REAL achieves an average success rate of 91.1\%.}
In complementary evaluation on 42 tasks from the RoboDojo benchmark, it achieves an average success rate of 84.4\%.
This study provides empirical guidance for turning physical exploration
into reusable robot skills that support continued improvement and
robust, efficient deployment. 
}
\checkdata[Project Page]{\url{https://dagroup-pku.github.io/PyRUA-Lean/real/}}
\checkdata[Keywords]{Agentic Robot, Embodied RSI, Code as Policies}

\begin{document}
\maketitle

\section{Introduction}

\begin{figure}[!t]
    \centering
    \includegraphics[width=\linewidth]{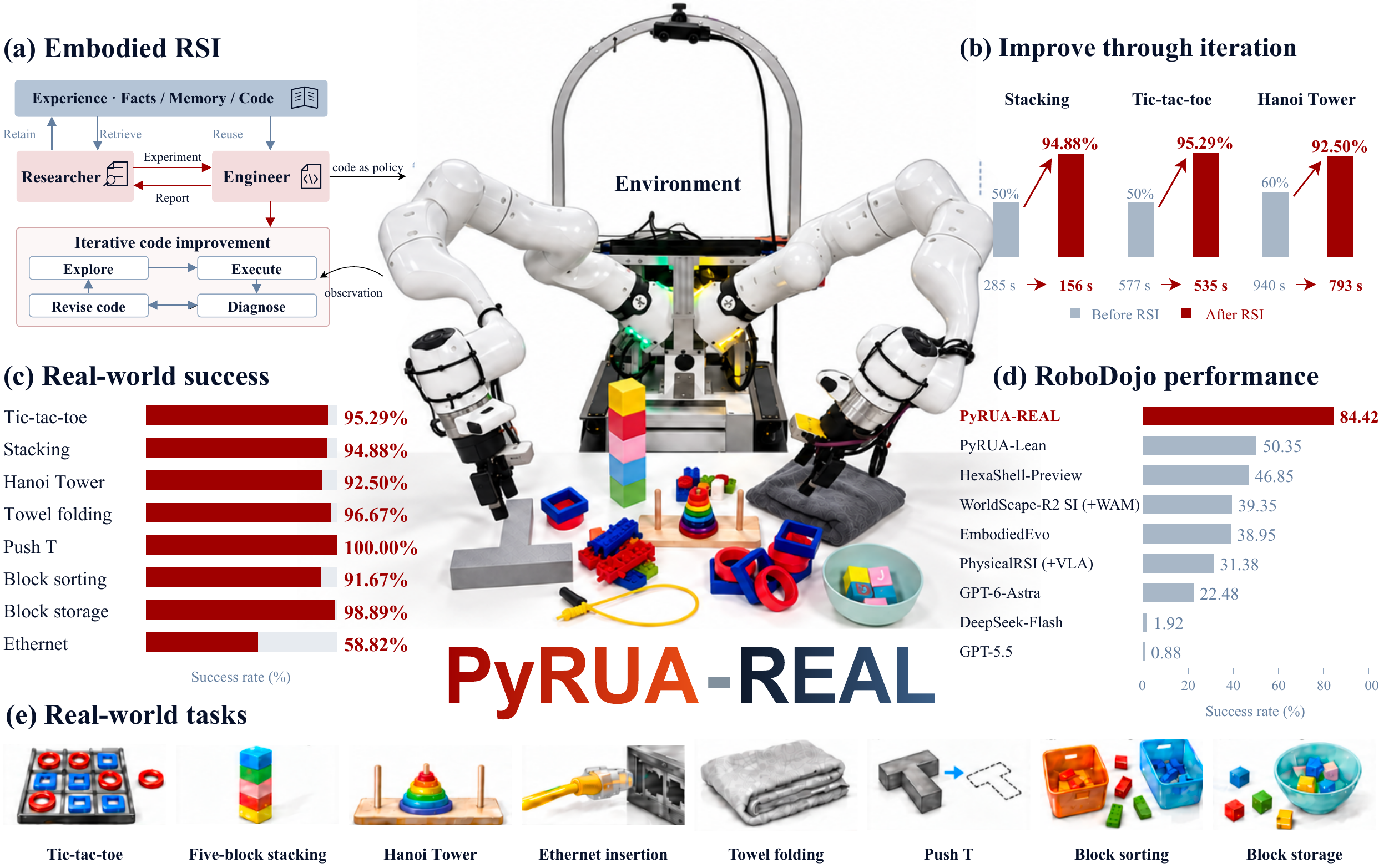}
    \caption{\textbf{PyRUA-REAL: Python-based Robot-Use Agents for Robot Exploration and Adaptive Learning.} (a)  Coding agents iteratively explore, execute,
    diagnose, and revise robot programs, retaining rollout experience
    as persistent memory and reusable skills. (b) Development-stage improvements in success rate and cycle time, including reset, for stacking, tic-tac-toe, and Hanoi. (c) Real-world success rates aggregated over development attempts for seven agent-free skills. (d) RoboDojo performance compared with eight baselines, measured by success rate averaged equally over five capability groups. (e) The eight real-world tasks include semantic reasoning, precise manipulation, contact-rich manipulation, deformable object manipulation and long-horizon tasks.}
    \label{fig:teaser}
\end{figure}

Large language models can translate task instructions into robot programs
that combine perception, spatial reasoning, and feedback
control~\cite{liang2023code,huang2023voxposer}. Coding agents powered by
models such as GPT-6-Astra extend this interaction by executing code,
inspecting observations, and adapting subsequent actions through execution
feedback~\cite{si2026fewer}. These capabilities make code a practical
interface between high-level reasoning and physical behavior. They also
raise a question beyond completing an individual task:
\emph{How can coding agents turn physical interaction experience into
robot capabilities that persist and improve across successive attempts?}

Experience-based adaptation and iterative code modification offer ways
to improve agents while keeping foundation model parameters
fixed~\cite{shinn2023reflexion,zelikman2023stop}. In robotics, recent work
has explored skill and agent-harness evolution in
simulation~\cite{wang2026shaper}, simulation-based practice followed by
physical deployment~\cite{wang2026rpg}, and policy improvement through
direct robot interaction~\cite{xiao2026enpire,wang2026sear}. Building on
these developments, we study embodied recursive self-improvement (RSI)
as a process in which execution experience informs code revisions and is
retained as memory and reusable skills for subsequent exploration.
Our focus is an empirical account of this process in \emph{real-world}
manipulation, tracing how capabilities are acquired, refined beyond first
success, and converted into programs for deployment.

Physical exploration presents challenges at several levels. Agents must
decide which experiments to run, which implementation details to revise,
and when to reconsider the underlying approach. Repeated local changes
may leave an unsuitable perception or control strategy intact. Execution
feedback is also ambiguous: an apparent task failure can originate in
motion, perception, or the procedure used to verify success. Moreover,
success under one configuration does not establish robustness to changes
in object positions, orientations, or task requirements. Retaining a
successful implementation therefore raises further questions about its
reuse on new tasks and its ability to operate without repeated online
agent decisions. Understanding embodied RSI requires examining these
aspects of the improvement process alongside task success.

We present PyRUA-REAL (Python-based Robot-Use Agent for Robot Exploration and Adaptive Learning) as an embodied RSI framework for this study (Figure~\ref{fig:teaser}). A researcher
agent selects experimental objectives and technical approaches, while an
engineer agent implements programs, executes robot trials, and diagnoses
failures. Their interaction connects local code improvement with task-level
reassessment. A persistent repository records environmental facts,
attempted approaches, execution outcomes, and reusable implementations,
allowing later experiments to build on earlier experience while model
parameters remain fixed. For deployment, the resulting code can be frozen
and executed without runtime language-model calls, while still responding
to live observations through its perception, control, and recovery routines.
This separation enables us to study both the process of developing robot
capabilities and the behavior of the programs that retain them.

Our experiments cover eight manipulation tasks on a dual-arm Franka
platform, spanning reasoning, long-horizon execution, precise contact,
deformable objects, and bimanual coordination. Recorded development success
averages 91.1\% across tasks. These rates aggregate attempts across code revisions; frozen-program
performance is evaluated separately. Complementary evaluation on 42 RoboDojo
tasks yields 84.82\% success, averaged equally over five capability groups,
compared with reported results of 50.35\% for PyRUA-Lean and 22.48\% for
GPT-6-Astra. Our main contributions can be summarized as:

\begin{itemize}
    \item \textbf{An empirical study of skill acquisition and
    continued improvement through real-world interaction.}
    Across eight physical manipulation tasks, we trace how coding
    agents turn exploration experience into executable robot skills
    and improve them beyond first successful rollout, documenting gains and
    trade-offs in task completion, execution time, and placement
    precision across program revisions.

    \item \textbf{Case studies on the factors shaping embodied RSI.}
    Through comparative experiments and diagnostic case studies,
    we examine how agent coordination, execution feedback, explicit
    task requirements, and previously acquired skills affect
    revision efficiency, robustness, and transfer to new tasks.

    \item \textbf{An evaluation of the resulting robot capabilities
    and their deployment costs.}
    Real-world evaluations compare frozen programs with online agent
    execution, quantifying execution-time benefits and identifying
    situations that still require online assistance.
    Complementary evaluation on 42 RoboDojo tasks measures performance
    across five capability groups against reported baselines.
\end{itemize}

\section{Related Work} \label{Sec: related works}
Our work lies at the intersection of programmatic robotic control and agent-driven self-improvement, aiming to enable robots to continuously improve their capabilities through real-world interaction. We review related research in LLM for robotic control, agentic RSI, and embodied RSI.

\textbf{LLM for Robotic Control.}
Large language models (LLMs) have enabled robot control by composing perception, planning, and control primitives into executable programs~\cite{liang2023code,huang2023voxposer,huang2024rekep}. Recent advances further explore direct agent-driven control~\cite{jia2026agentaspolicy,zhang2026unexpected,sensenova2026embodied}, as well as code-based approaches that improve action abstraction, execution feedback, and runtime efficiency~\cite{fu2026capx,si2026fewer,elmaaroufi2026rho}. Complementary approaches integrate programmatic skill representations with learned world and action models~\cite{liu2026agenticwam}. Agents can also propose structural priors that shape skill policy training and inform skill composition~\cite{jiang2026appl}. While online agent control offers flexible reasoning and adaptation, frequent model inference introduces computational overhead and latency. In contrast, executable programs enable efficient closed-loop control without continuous LLM intervention. Building on this paradigm, we employ coding agents to construct and progressively refine both reusable robot primitives and task-specific policies, while delegating high-frequency control to independently executable robot programs.

\textbf{Agentic RSI.}
Recent research has explored improving LLM-based agents through accumulated experience and autonomous modification, without necessarily updating model parameters. Experience-driven approaches retain reflections, interaction histories, and reusable knowledge to improve subsequent decisions~\cite{shinn2023reflexion,zhao2023expel}. Beyond memory-based adaptation, agentic RSI methods optimize the agents' underlying scaffolds, execution logic, and search procedures through iterative code modification and evaluation~\cite{zelikman2023stop,hu2024automated,yin2024godel,zhang2025darwin}. These approaches demonstrate how improvements can persist beyond individual attempts and benefit future problem-solving. Inspired by these developments, our work extends agentic self-improvement to physical interaction, where coding agents continuously accumulate execution-grounded knowledge and reusable control capabilities to improve subsequent robot policy generation and refinement.

\textbf{Embodied Recursive Self-Improvement.}
Recent embodied RSI research leverages environmental feedback to autonomously improve robot policies, reusable skills, and agent capabilities. Simulation-centered approaches evolve skills and agent harnesses in virtual environments, sometimes followed by real-world transfer~\cite{wang2026shaper,wang2026rpg,song2026embodiedrsi}. Other works investigate autonomous policy improvement through direct physical experimentation~\cite{xiao2026enpire,wang2026sear}, as well as reusable program libraries, hierarchical skill refinement, and persistent experience accumulation~\cite{lu2026aspire,wen2026roborsi,hu2026embodiedturing,li2026embodiedevo}. Building on these advances, our work emphasizes the complete process of exploration, improvement, and deployment in the real world, rather than treating physical robots merely as a final validation platform. 

\section{Methodology} \label{Sec: dataset}

\subsection{Problem Formulation}

We study how executable robot policies improve through environment interaction while foundation model parameters remain fixed. Improvements persist as reusable skills and experience. Given a task $\tau \sim \mathcal{D}$, the system receives observation $o_t$ at time $t$ and generates an action using a task-specific program $\pi_\tau$ and a shared skill library $\mathcal{L}$:
\begin{equation}
    a_t = \pi_\tau(o_{\leq t}; \mathcal{L}).
\end{equation}
The program may incorporate perception, planning and feedback control. A persistent experience repository $\mathcal{M}$ stores previous programs, experimental results, failure analyses, and their applicability conditions to support subsequent improvement.

At improvement iteration $k$, the system executes the current program, collects evidence $E^k$, and updates its policy, skills, or experience through an improvement process $\mathcal{U}_\theta$:
\begin{equation}
\begin{aligned}
    &(\pi_{\tau}^{k+1}, \mathcal{L}^{k+1}, \mathcal{M}^{k+1}) \\
    &\quad =
    \mathcal{U}_\theta
    (\tau, \pi_{\tau}^k, \mathcal{L}^k, \mathcal{M}^k, E^k),
\end{aligned}
\end{equation}
where $\theta$ denotes the fixed foundation model parameters. Execution feedback guides program revisions, which are evaluated in subsequent tests.

Let $J_\tau(\pi)=\mathbb{E}_{\xi\sim P_\tau^\pi}[R_\tau(\xi)]$ denote expected policy performance, where $\xi$ is an execution trajectory and $R_\tau$ is a predefined evaluation function. The objective is to obtain effective policies within interaction and computational budgets $B_\mathrm{int}$  and $B_\mathrm{comp}$:
\begin{equation}
\begin{aligned}
    \max_{\mathcal{U}_\theta}\quad
    & \mathbb{E}_{\tau\sim\mathcal{D}}
      \left[J_\tau(\pi_\tau^{B})\right], \\
    \text{s.t.}\quad
    & C_{\mathrm{int}} \leq B_{\mathrm{int}}, \qquad
      C_{\mathrm{comp}} \leq B_{\mathrm{comp}},
\end{aligned}
\end{equation}
where $\pi_\tau^{B}$ is the final policy obtained within the budgets, and $C_{\mathrm{int}}$ and $C_{\mathrm{comp}}$ denote interaction and computational costs during development.

We examine within-task policy improvement and cross-task experience reuse in terms of policy acquisition cost, final performance, and execution cost. The accumulated $\mathcal{L}$ and $\mathcal{M}$ can support reuse across tasks.

\begin{figure}[t]
    \centering
    \includegraphics[width=\linewidth]{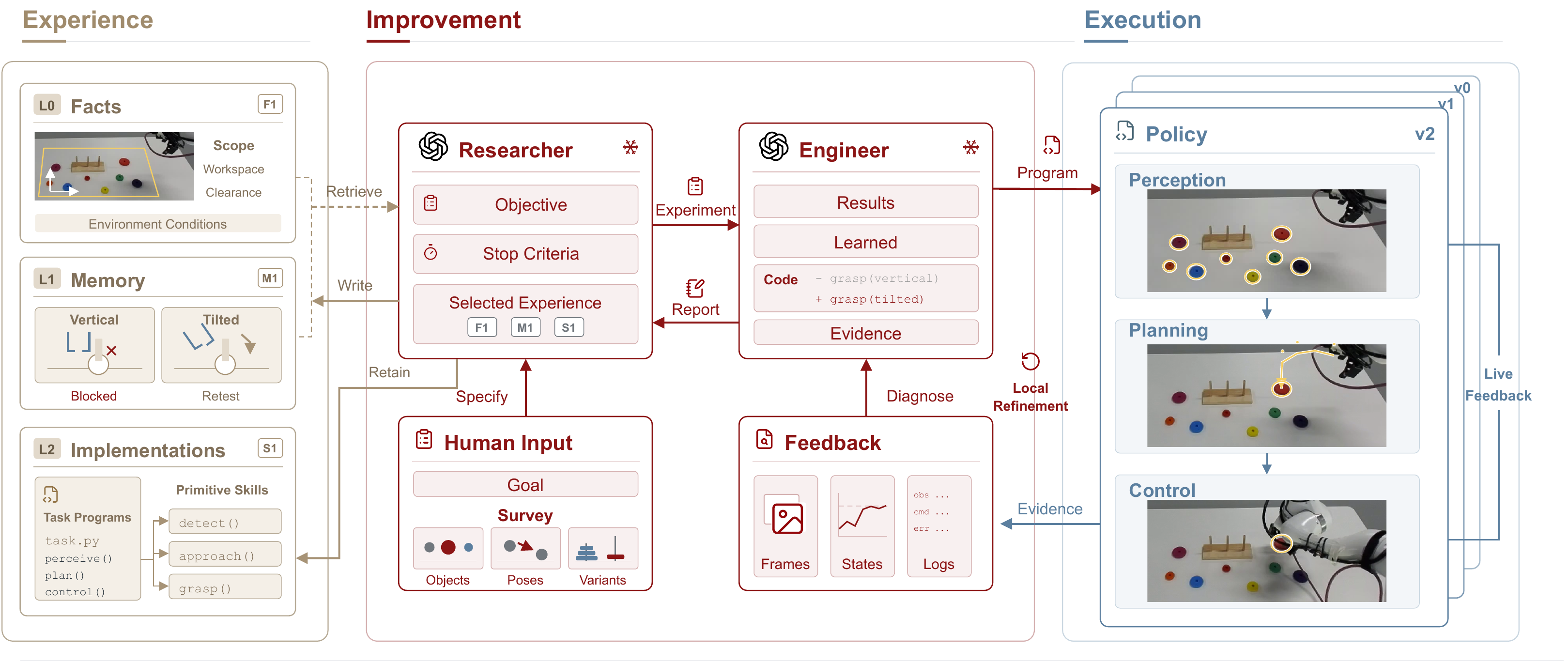}
    \caption{PyRUA-REAL develops executable policies through researcher--engineer collaboration and retains facts, memories, and code for reuse. The resulting policy uses live feedback during execution.}
    \label{fig:system_overview}
\end{figure}

\subsection{System Overview}
\label{sec:system_overview}

PyRUA-REAL consists of three components: \emph{improvement}, \emph{execution}, and \emph{experience}, as illustrated in Figure~\ref{fig:system_overview}. The system receives a human-provided task goal (\emph{goal}) and a task specification (\emph{survey}), which describes the required coverage of objects, positions, orientations, and task variants. With the underlying agent model parameters held fixed, the system develops task programs through code generation, testing, and feedback analysis. The resulting task program runs independently of the development agents.

The improvement component comprises two agents with complementary responsibilities. The \emph{researcher} manages the overall technical approach and specifies the objective, stopping criteria, and relevant experience for each experiment. The \emph{engineer} implements these requirements, diagnoses problems through environmental interaction, and revises the code. They coordinate through structured \emph{experiment} specifications and \emph{report} artifacts, connecting task-level experimental decisions with implementation-level exploration.

The execution component independently runs the task program, whose perception, planning, and control components use live observations to generate actions and advance the task. The program may invoke locally deployed pretrained models, such as SigLIP \cite{zhai2023sigmoidlosslanguageimage} and SAM\cite{kirillov2023segment}, as perception components. Execution frames, states, and logs support subsequent program revisions. The researcher organizes the resulting environmental facts, problem-solving records, and reusable implementations in the experience repository for future experiments and tasks.

\subsection{Coding Agents Collaboration}
\label{sec:intra_task_improvement}

The \textit{researcher} and \textit{engineer} address problems at different levels of granularity. The \textit{researcher} considers the complete task and its required variations to choose a technical approach and decide when to change methods or advance to the next stage. The \textit{engineer} focuses on implementation, environment interaction, and local diagnosis within the current experiment. Their basic unit of coordination is an \emph{experiment}: a goal-directed exploration process that may include multiple code revisions and tests.

Each experiment specifies an objective, stopping criteria, and selected experience. The researcher defines the objective for the current stage, provides relevant facts, historical approaches, and skill references, and autonomously sets the stopping criteria. Besides goal completion, stopping conditions may include a rejected hypothesis, a bottleneck requiring a different technical approach, or repeated revisions without progress. When a stopping condition is met, the engineer hands the problem back to the Researcher for reassessment.

Within an experiment, the engineer may revise code and conduct multiple focused tests without executing the complete task each time. For example, when object recognition is unreliable at a particular step, the engineer can compare recognition methods at that stage. The engineer independently decides how to implement and validate these changes within the scope of the experiment.

At the end of the experiment, the engineer returns a \emph{report} containing the outcome and termination reason (\emph{results}), findings from the exploration (\emph{learned}), supporting execution records (\emph{evidence}), and the final implementation (\emph{code}). The researcher uses this report to decide whether to refine the current approach, change direction, or advance to another task stage and defines the next experiment accordingly.

\subsection{Persistent Experience and Skill Evolution}
\label{sec:cross_task_improvement}

\begin{table}
    \centering
    \small
    \caption{Experience hierarchy and its primary use by the two agents.}
    \label{tab:experience_hierarchy}
    \setlength{\tabcolsep}{4pt}
    \renewcommand{\arraystretch}{1.15}
    \begin{tabularx}{\linewidth}{
        @{}l
        >{\raggedright\arraybackslash}X
        >{\raggedright\arraybackslash}X@{}
    }
        \toprule
        Level & Stored content & Primary use \\
        \midrule
        L0 (Facts)
        & Robot, workspace, and operating conditions.
        & Researcher: understand relevant environmental conditions. \\
        \addlinespace
        L1 (Memory)
        & Problems, attempted approaches, outcomes, and failure causes.
        & Researcher: identify relevant historical tasks, approaches, and skills. \\
        \addlinespace
        L2 (Implementation)
        & Task programs and reusable primitives.
        & Engineer: inspect, adapt, and test implementations. \\
        \bottomrule
    \end{tabularx}
\end{table}

Experience is organized into three levels, connecting environmental knowledge and problem-solving records to reusable code (Table~\ref{tab:experience_hierarchy}).
Experience retrieval follows the agents' division of responsibility. When exploration begins or the engineer reports an unresolved problem, the researcher consults \emph{facts} and \emph{memory} to select relevant historical tasks and skills. The engineer follows these references to inspect the code, assess its applicability, and adapt it to the current problem. Complete task implementations show how an approach is used throughout a task, while primitive functions provide smaller units for reuse. Adapted code must still be tested in the current environment.

The researcher updates the experience repository from the engineer's reports, preserving useful findings from both successful and unsuccessful attempts and linking them to the corresponding implementations. Skill usage is recorded based on the skills adopted in each experiment's final implementation; intermediate revisions and repeated calls do not count as additional uses. The researcher also decides which adapted implementations to retain as new skills and records their observed applicability and limitations.

\section{Experiments and Empirical Findings}
\label{sec:experiments}

In this section, we investigate how coding agents turn physical interaction experience
into improved, reusable robot programs. We first introduce the research
question and hypotheses (Section~\ref{sec:research_question}), describe
the experimental settings (Section~\ref{sec:experimental_settings}), and
report capability acquisition and continued improvement beyond first
success (Section~\ref{sec:main_results}). We then examine agent coordination
(Section~\ref{sec:agent_coordination}), execution feedback
(Section~\ref{sec:execution_evidence}), and explicit task requirements
(Section~\ref{sec:generalization_requirements}), followed by cross-task
primitive-skill transfer (Section~\ref{sec:primitive_transfer}) and deployment
efficiency (Section~\ref{sec:execution_efficiency}). Real-world development
histories and diagnostic cases provide the primary empirical evidence,
complemented by simulation evaluations and comparisons.

\subsection{Research Question and Hypotheses}
\label{sec:research_question}

Our experiments address one overarching research question:

\medskip
\noindent
\emph{How should coding agents turn physical interaction experience into progressively more capable robot programs? }

Specifically, we examine the following hypotheses, which the experimental
results may support or challenge:

\begin{itemize}

    \item \textbf{H1: Accumulated physical experience supports
    sustained self-improvement.}
    Retaining execution experience as memory and executable
    skills enables successive improvement cycles to build on
    prior discoveries and failures, improving robot capabilities
    beyond the first successful program. (Section~\ref{sec:main_results})

    \item \textbf{H2: Multi-agent coordination with specialized
    responsibilities improves RSI efficiency.}
    Distributing task-level planning, implementation, and independent
    review across specialized agents is expected to improve the
    efficiency of successive exploration and revision cycles. (Section~\ref{sec:agent_coordination})
    
    \item \textbf{H3: Rich execution feedback improves
    RSI efficiency.}
    Visual records and robot-state information enable more
    effective failure diagnosis and program revision than
    outcome-only feedback. (Section~\ref{sec:execution_evidence})
    
    \item \textbf{H4: Explicit task requirements improve
    the robustness of robot programs.}
    Specifying objects, layouts, and task variants
    guides exploration and skill refinement toward better
    robustness on unseen task variations. (Section~\ref{sec:generalization_requirements})

    \item \textbf{H5: Developed primitive skills can positively transfer to new tasks.} Agents given a primitive-skill library acquired on earlier tasks are expected to develop more successful programs for new tasks than agents without that library. (Section~\ref{sec:primitive_transfer})

    \item \textbf{H6: Consolidating exploration into executable programs improves deployment efficiency.}
    Encoding task decisions in reusable programs reduces repeated online agent calls and execution time. (Section~\ref{sec:execution_efficiency})

\end{itemize}

\subsection{Robotic Platforms and Task Settings} 
\label{sec:experimental_settings}

\begin{figure}[t]
    \centering
    \includegraphics[width=\linewidth]{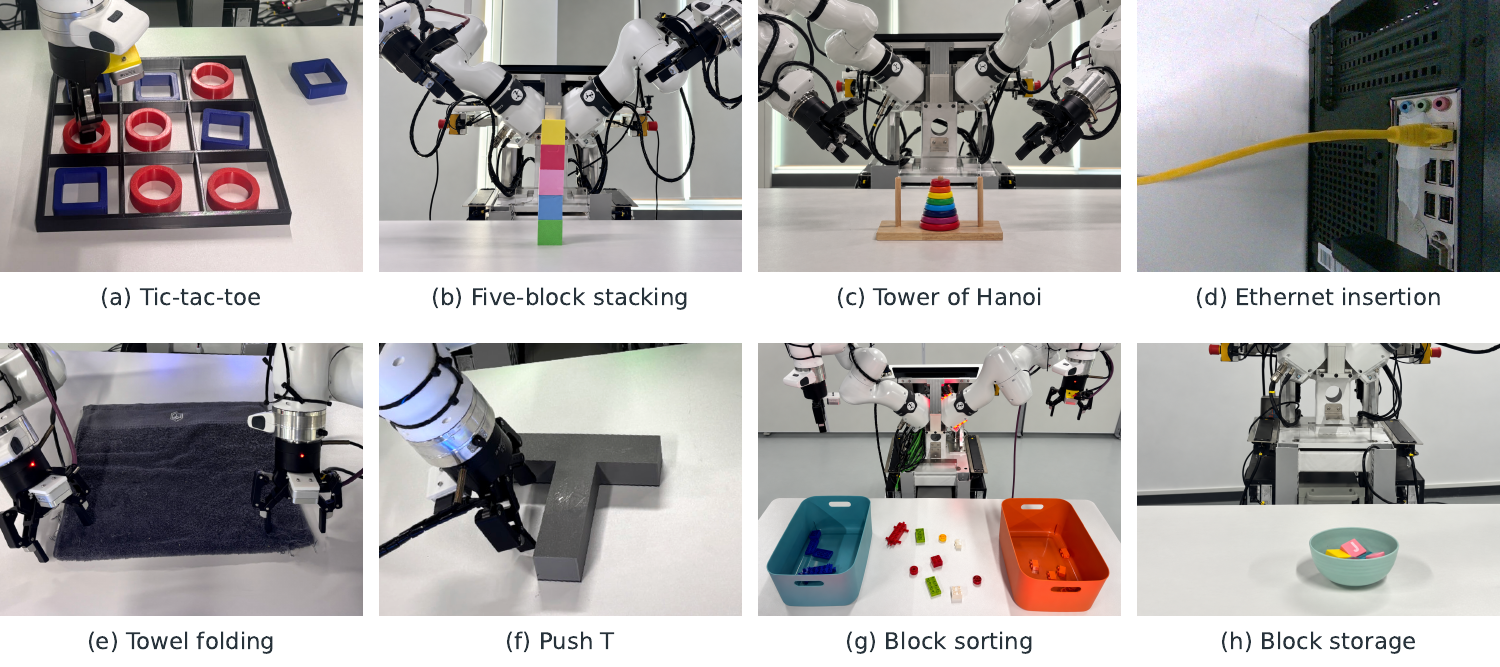}

    \caption{Eight real-world tasks. Top row: (a) tic-tac-toe, (b) five-block stacking, (c) Tower of Hanoi, and (d) Ethernet insertion. Bottom row: (e) towel folding, (f) Push T, (g) block sorting, and (h) block storage. Images show scenes recorded during task execution.}

    \label{fig:real_world_tasks}
\end{figure}

\begin{figure}[h]
    \centering
    \includegraphics[width=\linewidth]
        {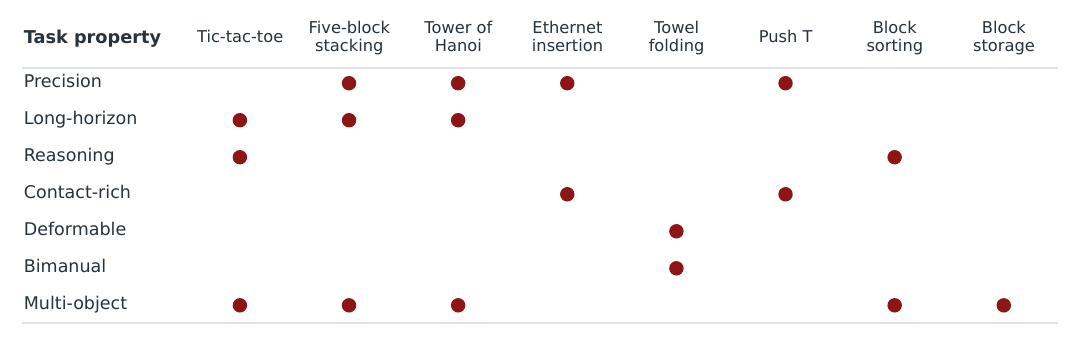}
    \caption{Selected properties of the eight physical tasks on the
        dual-arm Franka platform: precision, long-horizon execution,
        reasoning, contact-rich interaction, deformable-object handling,
        bimanual coordination, and multi-object handling. Marks indicate
        selected task properties, rather than measured difficulty or
        performance; unmarked cells do not imply that a property is absent.}
    \label{fig:task_properties}
\end{figure}

We evaluate PyRUA-REAL on a dual-arm Franka platform across
eight physical tasks, complemented by RoboDojo simulation
tasks. Figure~\ref{fig:real_world_tasks} shows the physical task
scenes, and Figure~\ref{fig:task_properties} summarizes the
selected properties of the physical tasks.

\paragraph{Real-world robot platform.}
The system comprises two Franka Research 3 arms with
Robotiq-85 grippers, a head-mounted ZED Mini camera, and
wrist-mounted Intel RealSense D405 cameras. Agents receive
interfaces for accessing sensor observations and issuing
low-level control commands. Platform calibration results
and task-specific controllers and solvers are not provided
in advance; they must be established during exploration.

\paragraph{Real-world tasks.}
We consider eight different real-world tasks illustrated in Figure~\ref{fig:real_world_tasks} with evaluation focuses summarized in Figure~\ref{fig:task_properties}: 
\textbf{Tic-tac-toe}, alternating legal piece
placements on a $3\times3$ board;
\textbf{Five-block stacking}, constructing a tower in a
prescribed color order;
\textbf{Tower of Hanoi}, sorting eight disks by size and
placing them on a peg;
\textbf{Ethernet insertion}, inserting an RJ45 connector
into a fixed port;
\textbf{Towel folding}, folding a towel with both arms;
\textbf{Push T}, pushing a T-shaped object to a target pose;
\textbf{Block sorting}, placing blocks into color-matched
bins; and
\textbf{Block storage}, collecting scattered blocks into
a bowl.
Detailed task configurations and completion criteria are
provided in Appendix~\ref{app:task_settings}.

\paragraph{Simulation tasks.}
RoboDojo comprises 42 base tasks in five capability groups:
\textbf{Generalization},
\textbf{Memory},
\textbf{Precision manipulation},
\textbf{Long-horizon tasks}, and
\textbf{Open tasks}.
These groups cover object and layout variations, state and
sequence memory, precise alignment, multi-stage execution,
and instruction grounding and reasoning, respectively.
We use the full suite for capability evaluation and task
subsets for studies of execution feedback and explicit task
requirements; their evaluation protocols are specified with each study.

\subsection{Embodied RSI Performance in the Real World and Simulated Environments}
\label{sec:main_results}

We examine whether \textbf{physical interaction experience can be translated into executable robot capabilities and further improvements through embodied RSI (H1).} We first report results across the eight
physical tasks, then examine continued improvement beyond first successful exploration rollout and the limits of agent-free execution. Frozen-program evaluation on RoboDojo provides complementary evidence on a broader task suite.

\subsubsection{Real-world capability  acquisition with agent}
\label{sec:physical_acquisition}

Physical exploration produces agent-free programs for seven of the
eight tasks (Figure~\ref{fig:physical_results}). These programs execute
using live perception, control, and recovery logic without runtime
language-model calls. Approximate time to first success ranges from
1.5 hours for block storage to 45 hours for Push T, indicating substantial
variation in acquisition cost across tasks. Ethernet insertion remains
unsuccessful without online agent participation in the reported
development record.

Across the seven tasks with agent-free solutions, recorded development
success ranges from 91.67\% to 100\%
(Figure~\ref{fig:physical_results}b). These rates aggregate attempts
across program revisions and describe the development record rather
than the reliability of a single frozen program. Successful execution
times also vary substantially across tasks
(Figure~\ref{fig:physical_results}c); their task-specific endpoints
are defined in Appendix~\ref{app:task_settings}.

\begin{figure}[t]
    \centering
    \includegraphics[width=0.98\linewidth]{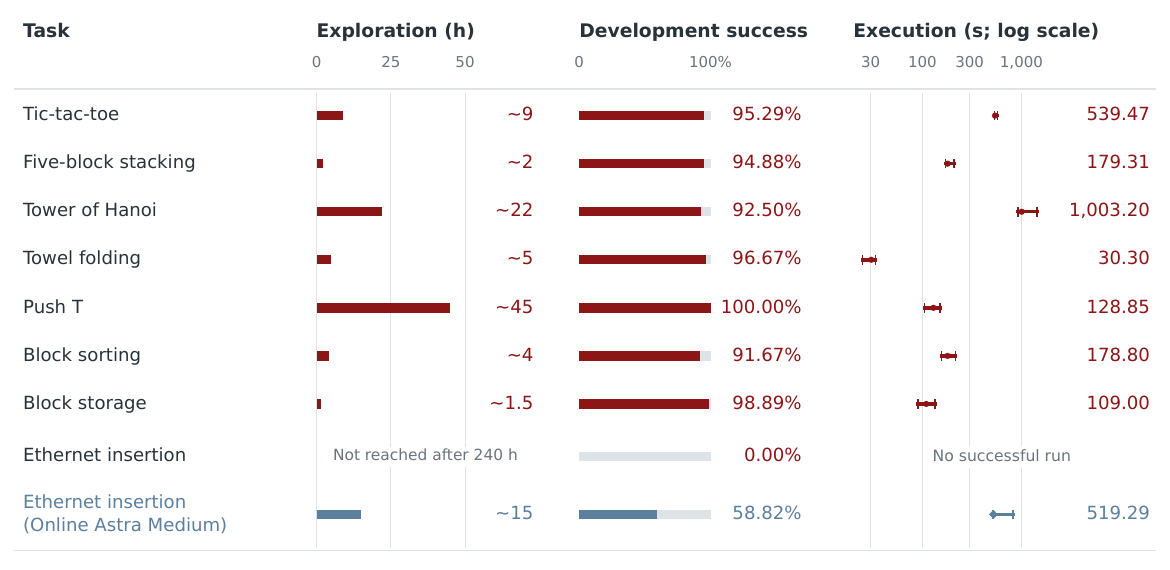}
    \caption{
    \textbf{Real-world capability acquisition with agent.}
    The first column is approximate time to first complete success.
    Agent-free Ethernet insertion remains unsuccessful after ten days;
    online assistance yields success after approximately 15 additional
    hours of exploration.
    The
    second column recorded success rates across development attempts and
    program revisions. Execution-time points show successful-run medians; whiskers span
    the 25th to 75th percentiles. Only the execution-time axis is logarithmic.
    }
    \label{fig:physical_results}
\end{figure}

\subsubsection{RSI beyond the first successful rollout}
\label{sec:continued_physical_improvement}

The first successful execution establishes that a task is feasible, but
leaves room to improve repeatability, speed, and placement precision.
To examine H1, we analyze the subsequent development histories of stacking,
tic-tac-toe, and Hanoi. Successive program versions show further gains,
with occasional regressions and different trajectories across tasks.
Figure~\ref{fig:physical_progress} compares completion and execution time across the three tasks; Figure~\ref{fig:block_placement} shows stacking placement errors in position and angle. The results presented here are from exploring process and do not represent the final results of repeat tests.

\begin{figure}[t]
    \centering
    \includegraphics[width=0.95\linewidth]{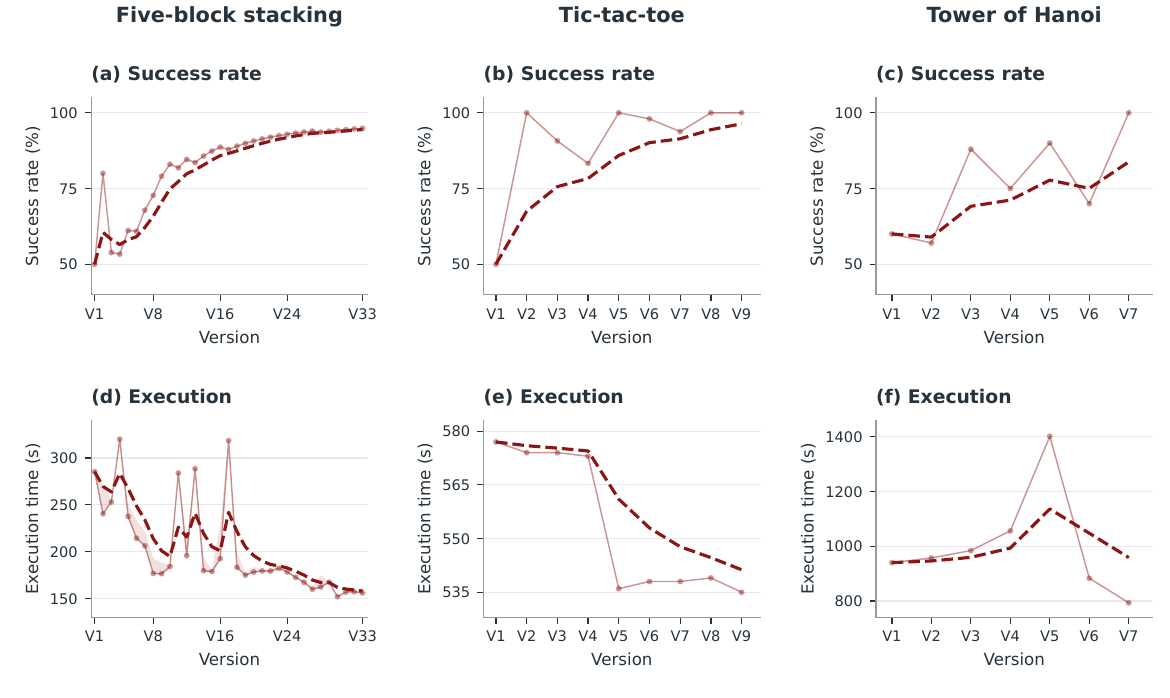}
    \caption{\textbf{Continued improvement after first success.}
    Columns show five-block stacking, tic-tac-toe, and Hanoi.
    (a--c) Success rate; (d--f) execution time, including reset.
    Shading in (d) shows the interquartile range.
    Points are observed values; dashed curves are exponential moving averages
    ($\alpha=0.35$). }
    \label{fig:physical_progress}
\end{figure}

Stacking improves gradually across revisions.
Repeated refinement of the transfer routine improves both execution time and placement estimates. Median stack--unstack cycle time falls from 285.3 to 156.0\,s, while mean visually estimated lateral error decreases from 2.69 to 0.44\,mm for stacking
(Figures~\ref{fig:physical_progress} d and \ref{fig:block_placement} a). Full-cycle success rises from 88.7\% to 98.7\% across the two reported smooth-stage evaluations; these stage-level rates are distinct from the cumulative success curve in the figure. Despite local regressions, the reductions in time and estimated error extend across multiple revisions, consistent with incremental improvements to a repeatedly executed routine. Mean absolute angular error for stacking also falls from 1.37 to 0.25$^\circ$ (Figure~\ref{fig:block_placement}b). These errors are visual estimates, rather than independent measurements of physical accuracy.

\begin{figure}[h]
    \centering
    \includegraphics[width=0.95\linewidth]{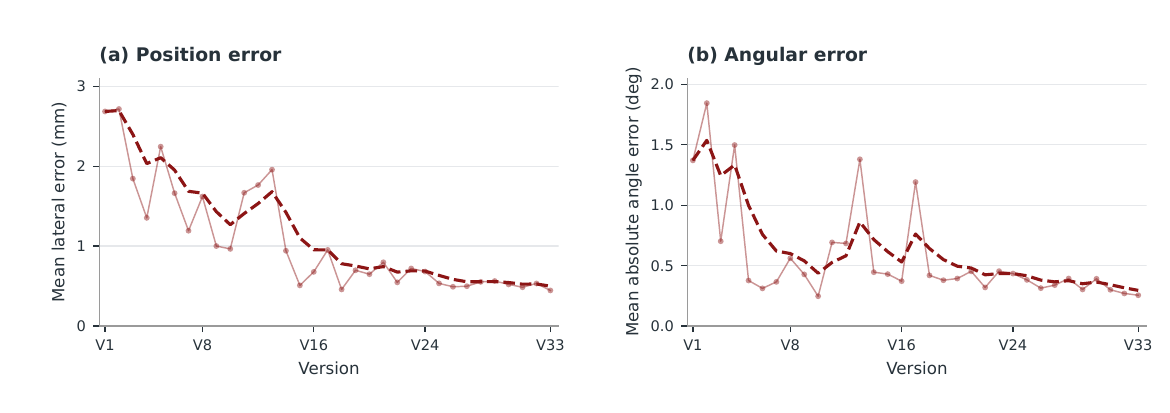}
    \caption{\textbf{Five-block placement precision across revisions.}
    (a) Mean lateral error and (b) mean absolute angular error. Both panels use the same placements
    from completed cycles in the 33 batches shown in
    Figure~\ref{fig:physical_progress}a,d. Points show batch means; dashed
    curves are exponential moving averages ($\alpha=0.35$).}
    \label{fig:block_placement}
\end{figure}

Tic-tac-toe shows a significant change after a motion-strategy revision.
Success rate rises from 50\% in the first version grow to 100\%
in the next, with later regressions before returning to 100\%.
Game-and-reset time falls from 577 to 535\,s overall
(Figure~\ref{fig:physical_progress} b,e). Most of this reduction occurs
between V4 and V5, when time drops from 573 to 536\,s. At this transition,
a revision replaces sequences of elevated free-space waypoints with a
single \texttt{JointMotion} for transfers repeated during play and reset.
The two relatively stable timing ranges on either side of this revision
link the observed improvement to a change in motion strategy. The
success-rate regressions also show why faster execution must be
checked alongside task success.

Hanoi tower shifts from improving reliability to reducing execution time.
Early revisions prioritize reliable completion, with longer execution
times. Median stack--unstack cycle time reaches 1402\,s in the robust
stage, then falls to 883 and 793\,s in the two speed-focused versions
(Figure~\ref{fig:physical_progress} c,f). Full-cycle success rate fluctuates
between stages, rising from 60\% initially to 100\% in the final
version. This history illustrates a change in optimization priorities:
first establish a reliable routine, then reduce its execution cost.
Final version combines shorter execution with higher success rate, although intermediate revisions do not improve both together.

These histories show that embodied RSI can continue improving an executable
skill after the first successful trial. Improvement can accumulate through
local refinement or follow a change in motion strategy, and its direction
can shift as reliability and speed become different priorities.

\subsubsection{Additional Evaluation on RoboDojo}
\label{sec:main_simulation_results}

\begin{figure}[h]
    \centering
    \includegraphics[width=0.95\linewidth]{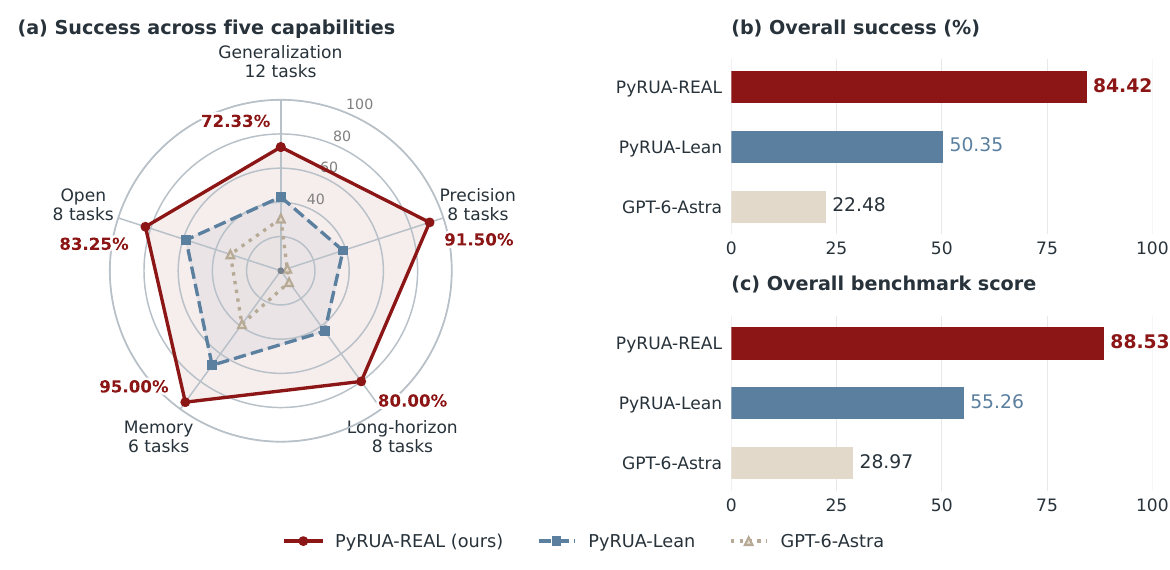}
    \caption{\textbf{Complementary RoboDojo evaluation.}
    Reported results for PyRUA-REAL, PyRUA-Lean, and GPT-6-Astra in the
    October 24, 2026 comparison snapshot: (a) success rates by capability,
    (b) overall success, and (c) benchmark score. Overall metrics weight
    the five groups equally; radar labels give task counts.
    PyRUA-Lean results are from its official evaluation. Full method
    comparisons, with column-best values in bold, appear in
    Appendix~\ref{app:simulation_results}.}
    \label{fig:robodojo_comparison}
\end{figure}

Across 42 RoboDojo tasks, PyRUA-REAL's frozen programs achieve 84.42\%
capability-averaged success, compared with 50.35\% for PyRUA-Lean and
22.48\% for GPT-6-Astra
(Figure~\ref{fig:robodojo_comparison}). The largest advantage over
PyRUA-Lean is in Precision (74.00\% versus 38.25\%), followed by
Long-Horizon (56.00\% versus 43.75\%), while Generalization remains
slightly lower (41.17\% versus 43.17\%). The programs therefore perform
better overall, with gains concentrated in precision manipulation and
long-horizon execution. These are reported results from separate
evaluations with different or unspecified seeds.

\FloatBarrier

\subsection{Agent Coordination for Exploration}
\label{sec:agent_coordination}

We examine H2 through physical development histories, asking whether
specialized agent responsibilities help programs improve through
successive physical trials. We analyze the cost of adding a reviewer
to the engineer's revision and testing cycle, then examine cases in
which local revisions continue without resolving a stalled approach.

\begin{figure}[t]
    \centering
    \includegraphics[width=0.95\linewidth]{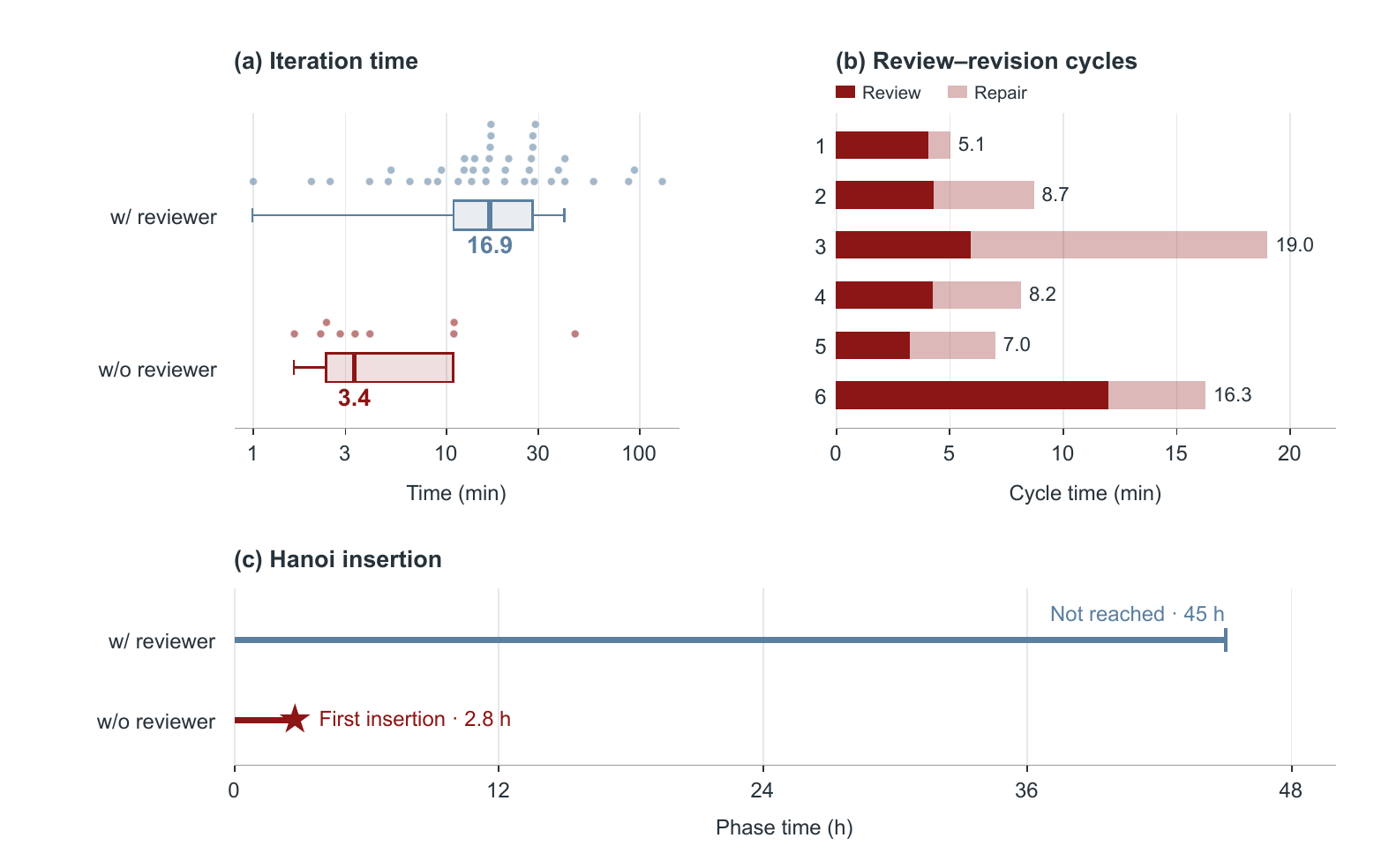}
    \caption{Coordination costs with and without a reviewer.
    (a) Result-to-next-trial intervals on a logarithmic scale; boxes show
    medians and interquartile ranges.
    (b) Six review–revision cycles before hardware execution in the Hanoi experiment. Review includes dispatch and review; Repair includes modification and resubmission. Bar-end labels show total cycle time.
    (c) Elapsed time in the reviewed phase and the subsequent direct-testing
    phase, which reaches the first verified disk-on-peg insertion.}
    \label{fig:collaboration_turnaround}
\end{figure}

\paragraph{An additional reviewer agent can slow down exploration.}
We compare development records with and without a reviewer that
inspects each submitted program before hardware execution. The
reviewer can reject a submission but cannot modify it; the engineer
must therefore revise and resubmit rejected programs. The median
interval from reporting a result to starting the next physical trial
is 16.9 minutes with the reviewer and 3.4 minutes without, across
40 and 9 recorded transitions, respectively
(Figure~\ref{fig:collaboration_turnaround}a). Six submission--rejection--revision
cycles in the Hanoi experiment take 5.1--19.0 minutes each, with a
median of 8.5 minutes (Figure~\ref{fig:collaboration_turnaround}b), showing
how review adds waiting before the next physical test. The reviewed
phase records no verified disk-on-peg insertion over 45.0 hours,
whereas the subsequent direct-testing phase reaches its first
insertion 2.8 hours after the first grasp test
(Figure~\ref{fig:collaboration_turnaround}c). Because the phases are sequential,
the later phase may benefit from earlier exploration, so the insertion
timeline does not isolate the effect of removing the reviewer.
The directly observed review delays nevertheless identify a coordination
cost: assigning approval to a separate agent can reduce the rate at
which the system obtains new physical evidence for improvement.

\begin{figure}[t]
    \centering
    \includegraphics[width=0.95\linewidth]{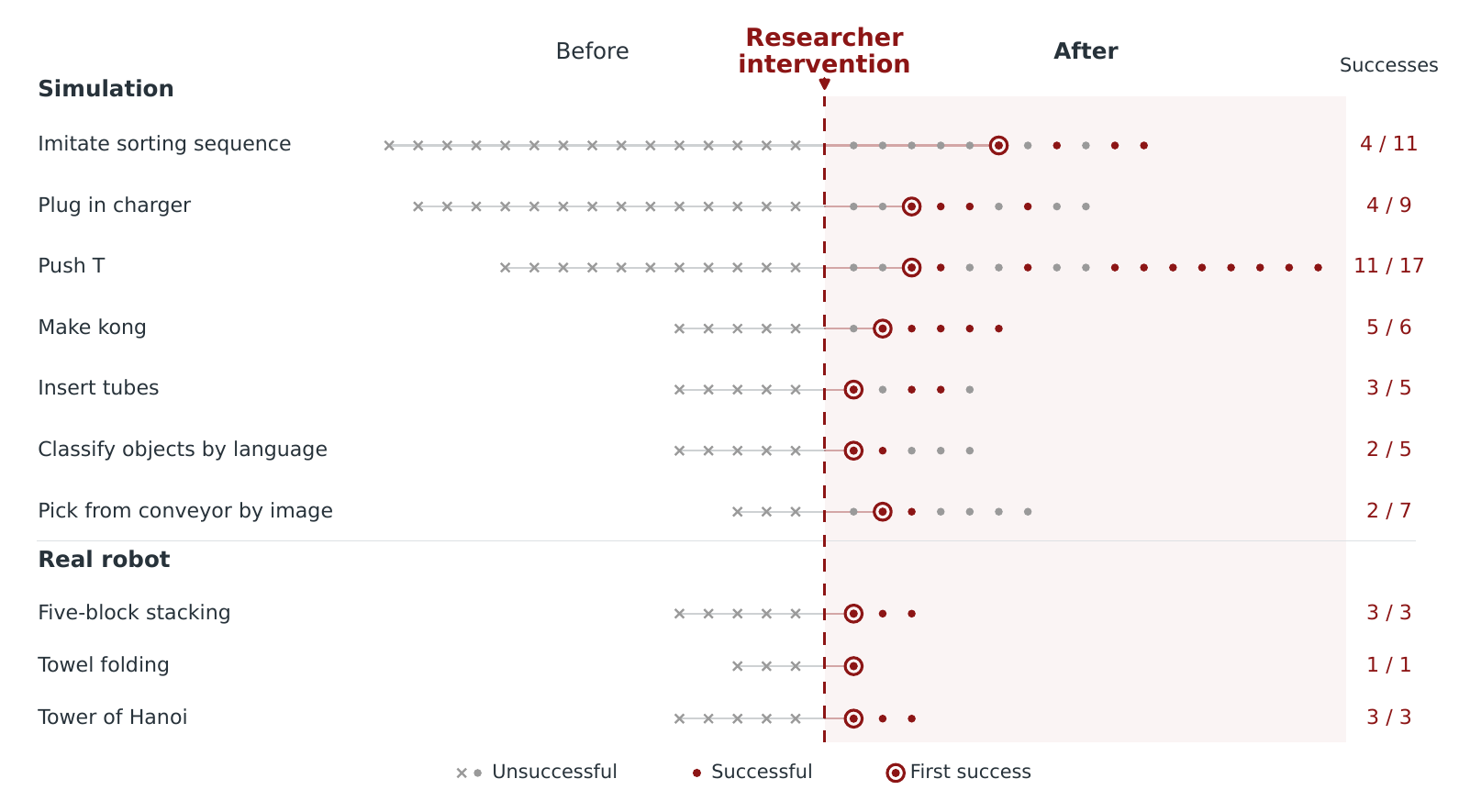}
    \caption{\textbf{Task-level reassessment following stalled local revisions.}
    Trial sequences for seven simulated tasks and three real-world tasks,
    aligned at reassessment by the Researcher (vertical dashed line).
    A single Researcher participates throughout each history; intervention
    denotes its reassessment of the current approach and redirection of
    exploration. Gray crosses and dots mark unsuccessful trials before
    and after reassessment, respectively; red dots mark successes, with
    circles highlighting the first success in each sequence. Right-hand
    fractions report successful trials over all recorded trials after
    reassessment. Horizontal spacing represents trial order, not elapsed
    time. Outcomes summarize development attempts across revisions rather
    than independent tests of a fixed program.}
    \label{fig:researcher_rollout_sequences}
\end{figure}

\paragraph{Task-level reassessment can break local revision deadlocks.}
Repeated local revisions can leave exploration stalled when the underlying approach remains unchanged. Figure~\ref{fig:researcher_rollout_sequences} shows seven simulated and three physical task histories in which unsuccessful trials are followed by success after task-level reassessment. A single Researcher is involved throughout each history; the intervention marks its reassessment of accumulated failures and redirection of the
Engineer's exploration. In all three physical cases, the next trial
succeeds, with subsequent development records totaling 3/3 successes for stacking, 1/1 for towel folding, and 3/3 for Hanoi. Progress in simulation is less uniform: Push T records 11/17 successes after reassessment, while Imitate sorting sequence records 4/11. These histories motivate bounding local revision and returning unresolved failures to the Researcher for reassessment. They associate reassessment with renewed progress, without isolating its effect from continued exploration.

\subsection{RSI Based on Execution Feedback}
\label{sec:execution_evidence}

We examine H3 by investigating how execution experience becomes
actionable evidence for embodied RSI. Physical diagnostic traces
connect inspected observations to program revisions and subsequent
validation, while a simulation study compares feedback interfaces
under a fixed revision budget.

\paragraph{Reliable improvement requires reliable execution feedback.}
Apparent execution failures can originate in perception or verification. In tic-tac-toe, merging two touching rings causes a false placement rejection; correcting the detector makes the same image pass the unchanged acceptance criterion, without moving the robot. In stacking, successful cycles still produce large reported block-return errors. Inspection reveals an unintended camera view; correcting view selection lowers the visual error estimates, but does not independently establish more accurate motion. Embodied RSI must therefore identify whether a discrepancy comes from execution or its measurement before revising control, perception, or verification.

\paragraph{Richer execution feedback improves task success with reduced optimization time.}

\begin{figure}
    \centering
    \includegraphics[width=\linewidth]{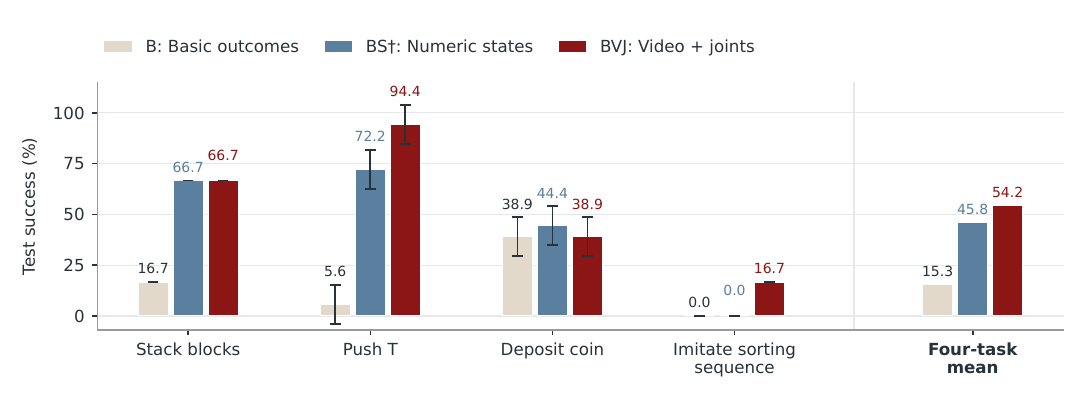}
    \caption{\textbf{Feedback interfaces affect program improvement.}
    Test success after five revisions with basic outcome reports (B),
    numeric state reports (BS$^{\dagger}$), or video and joint trajectories
    (BVJ). All conditions include success, termination reason, action
    count, and timing. BS additionally supplies sampled numeric robot
    and ground-truth object states to the revising agent
    ($\dagger$: privileged diagnostics); BVJ adds complete camera video
    and synchronized joint trajectories. Each task--condition pair has
    one development session and 18 test executions across six held-out
    layouts. The final group weights the four tasks equally.}
    \label{fig:feedback_results}
\end{figure}

We compare three feedback interfaces on four RoboDojo tasks:
basic outcome reports (B), numeric state reports (BS), and video
with joint trajectories (BVJ). BS includes ground-truth object states
as privileged diagnostics for revision; program execution inputs
remain identical across conditions (RGB-D, calibration, and robot state).
Each GPT-6-Astra Medium session generates an initial program and makes
five revisions on two development layouts. Final programs are evaluated
in 18 executions across six held-out layouts, with test outcomes
withheld from the revising agent. Mean success is 15.3\% for B,
45.8\% for BS, and 54.2\% for BVJ
(Figure~\ref{fig:feedback_results}). Compared with B, BS increases
success from 3/18 to 12/18 on Stack blocks, from 1/18 to 13/18 on
Push T, and from 7/18 to 8/18 on Deposit coin. Both B and BS record
0/18 on Imitate sorting sequence, while BVJ achieves 3/18.
The higher mean success with BVJ accompanies lower mean optimizer time
than B: 25.4 versus 31.8 minutes for initial generation and all five
revisions. Token usage is slightly higher, at 3.65 versus 3.54 million,
including cached inputs. Optimizer time covers agent turns and their
tools, excluding separate development and test evaluations.
These results show task-dependent benefits from richer revision
feedback; the BS condition additionally differs in access to privileged
object-state information. Each task--condition pair has one development
session, so variability across independent sessions remains unmeasured.
Appendix~\ref{app:feedback_analysis} provides the full eight-condition
protocol and cost analysis.

\FloatBarrier

\subsection{Towards More Robust Robot Programs}
\label{sec:generalization_requirements}

\begin{figure}[h]
    \centering
    \includegraphics[width=0.95\linewidth]{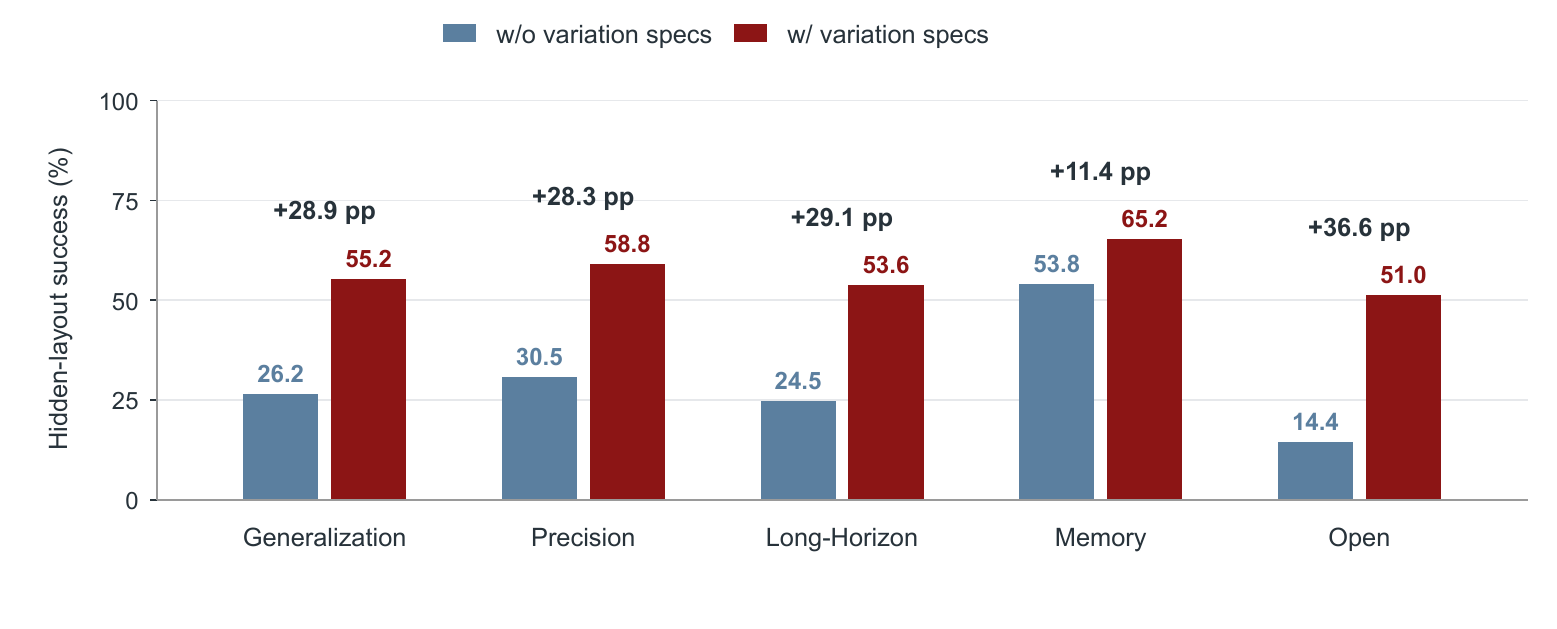}
    \caption{Success on held-out scenarios with and without continued
    exploration and program revision with explicit workspace boundaries
    and target position and orientation ranges. Both versions are evaluated
    on the same test scenarios. Category means weight tasks equally;
    annotations indicate absolute gains in percentage points.}
    \label{fig:generalization_survey}
\end{figure}

\paragraph{Robot programs become more robust to unseen scenarios when informed with explicit descriptions of the task variations (H4).}
Agents initially develop programs under an instruction to maximize success
on hidden test scenarios. We then provide numerical specifications and
images defining the workspace boundaries and the ranges of target positions
and orientations, and continue exploration and program revision.
We evaluate both program versions on the same held-out scenarios, measuring
robustness as success across these variations.
Mean success increases in all five RoboDojo categories, with gains of
11.4--36.6 percentage points (Figure~\ref{fig:generalization_survey}).
Open tasks improve from 14.4\% to 51.0\%, while Memory improves from
53.8\% to 65.2\%. At the task level, Stack bowls improves from 0.0\% to
86.4\% and Fasten screws from 9.1\% to 81.8\%, although gains are not
universal: Swap blocks declines from 68.2\% to 59.1\%.
The results suggest that specifying the operating conditions a program
must handle helps direct embodied RSI toward robustness across scenarios.
This comparison includes additional exploration, so it does not isolate
the effect of the added specifications alone.
Appendix~\ref{app:generalization} provides the evaluation protocol and
per-task results.

\FloatBarrier

\subsection{Cross-Task Experience Transfer}
\label{sec:primitive_transfer}
\begin{figure}[h]
    \centering
    \includegraphics[width=0.92\linewidth]{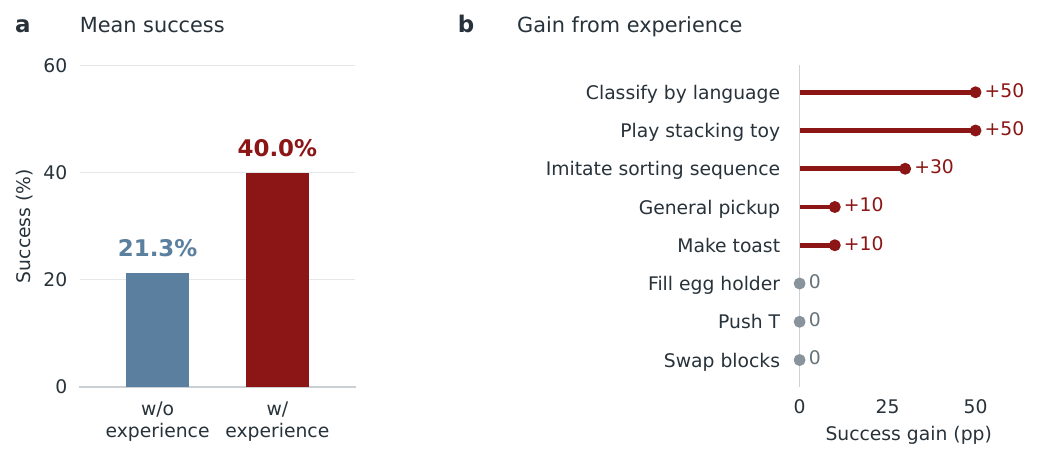}
    \caption{\textbf{Transfer from a primitive-skill library to new tasks.} Agents develop programs for eight target tasks with or without access to primitive skills acquired on earlier tasks. (a) Mean success rate across the target tasks. (b) Per-task success-rate differences, computed as with-library minus without-library performance. Red markers indicate positive gains; gray markers indicate no change. Five of the eight tasks benefit from the library.}
    \label{fig:primitive_skill_transfer}
\end{figure}

\paragraph{Previously developed primitive skills support positive transfer to new tasks(H5). }
In a separate eight-task comparison, agents start new-task exploration
with or without a primitive-skill library developed on earlier tasks.
Programs developed with the library achieve 40.0\% mean success, compared with 21.3\% without it (Figure~\ref{fig:primitive_skill_transfer}). Five tasks improve: classify by language and play stacking toy each gain 50 percentage points, Imitate sorting sequence gains 30, and general pickup
and make toast each gain 10. fill egg holder, push T, and swap blocks show no gain. These results indicate that skills retained from previous
experience can benefit subsequent tasks, with transfer benefits varying
across tasks.

\FloatBarrier

\subsection{Offline Coding: Execution Efficiency and Capability}
\label{sec:execution_efficiency}

\begin{figure}[H]
    \centering
    \includegraphics[width=0.92\linewidth]{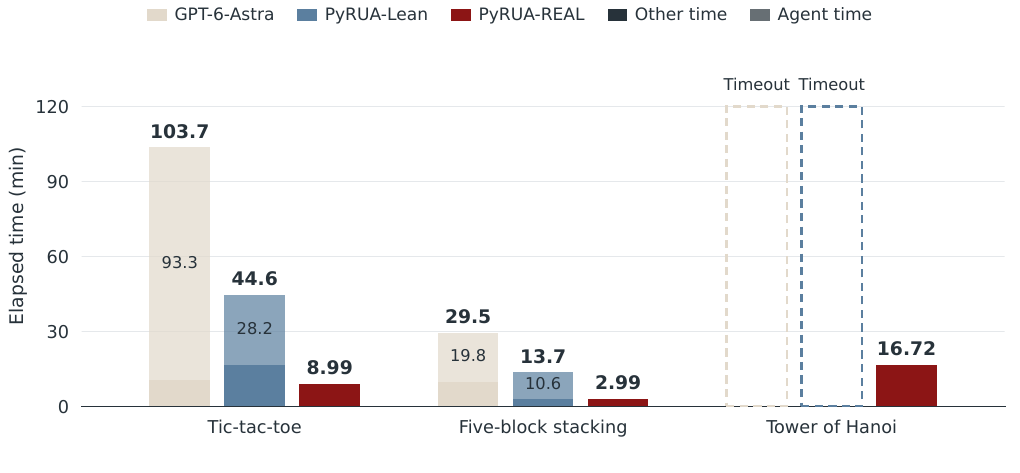}
    \caption{\textbf{Execution time and online thinking overhead.}
    Offline times are successful-run
    medians including reset; online bars show recorded times.
    Dashed outlines mark non-completion within 120 minutes with unmeasured
    timing splits. Development time is excluded.}
    \label{fig:real_execution_comparison}
\end{figure}

We distinguish two coding modes by when agent-generated code is produced
relative to deployment. In \textit{offline coding} (PyRUA-REAL), agents
develop and revise programs through physical exploration before deployment;
the resulting code is then frozen and executes without language-model calls.
Its sensing, conditional actions, and recovery logic still respond to live
robot observations. In \textit{deployment-time coding} (PyRUA-Lean), the agent writes
and executes code during the task, uses execution feedback, and generates
subsequent code as needed. A code segment can perform multiple actions and
local checks, but execution still includes agent calls between segments.
Figure~\ref{fig:real_execution_comparison} compares these modes with
\textit{online robot action inference without coding} , which selects
actions from visual observations and robot state without writing code or
accessing a shell.

\paragraph{Converting online exploration into fixed programs avoids repeated agent decisions(H6). }
In offline coding mode, the programs for tic-tac-toe, five-block stacking,
and Tower of Hanoi
complete execution and reset without runtime language-model calls, with
median successful cycle times of 8.99, 2.99, and 16.72 minutes, respectively.
On tic-tac-toe and stacking, online coding with PyRUA-Lean takes
approximately 4.6--5.0 times as long as offline coding with PyRUA-REAL;
online action selection with GPT-6-Astra takes approximately 9.9--11.5 times
as long (Figure~\ref{fig:real_execution_comparison}).
Agent thinking accounts for 63--90\% of the recorded online execution time.
For stacking, PyRUA-Lean spends 10.6 of 13.7 minutes thinking; its remaining
3.1 minutes are close to the program's 2.99-minute cycle. Thus, most of
this observed timing gap comes from waiting for agent decisions.
Neither online baseline completes Hanoi within 120 minutes.
The comparison concerns execution time and excludes the cost of online
exploration and program development.

\paragraph{Precise, contact-rich tasks can still require online agent assistance.}
The efficiency of offline coding depends on whether the frozen program
handles the conditions encountered during execution. Ethernet insertion,
which requires precise connector alignment and contact with the port,
exposes a limitation of the current code: the development records contain
0/4 successes with scripted execution and 10/17 with a workflow using
GPT-6-Astra Medium for online decisions. This assisted workflow combines
programmed motion with online assessment and decisions between stages;
it does not yet provide agent-free execution in offline coding mode.
In one successful run, motion takes 281.8\,s and inter-stage checks and
decision gaps take 220.8\,s, accounting for 44\% of the total 502.7\,s.
These gaps include visual checks and human review and therefore cannot
be attributed entirely to agent thinking.
The higher observed success rate supports retaining online assistance
where the current program remains unreliable. 

\FloatBarrier

\section{Discussion}
\label{sec:discussion}

\paragraph{Method selection can limit self-improvement.}
The deadlocks in Section~\ref{sec:agent_coordination} suggest that progress
can be limited by which methods an agent considers.
Depth and calibration errors constrained geometric localization: stacking
progressed with local visual alignment and contact feedback, while Hanoi
combined coarse localization with wrist-camera feedback.
In simulated general pickup, success improved after a human suggested
SigLIP, a visual model the agent had initially overlooked.
These cases indicate that the ability to implement and refine a method
does not ensure that useful alternatives will be explored.
A practical implication is to include explicit comparisons of alternative
perception and control strategies when the current approach repeatedly fails.

\paragraph{Human assistance can shape the skills agents acquire.}
The SigLIP case in the preceding paragraph illustrates how guidance can
expand the methods considered during exploration. Human input also
resolved an episode in which an agent repeatedly routed around a false
contact detection. When human assistance was unavailable, the Hanoi agent
developed a strategy for pushing hard-to-grasp disks into accessible
positions; this strategy was later reused in tic-tac-toe.
These observations suggest that intervention affects both immediate
progress and the recovery skills retained for future tasks.
Reports of autonomous improvement should therefore document what humans
contributed and which operations the resulting program performs independently.
The cases motivate studying when to intervene, without establishing an
optimal level of assistance.

\paragraph{Offline programs may still need online agent assistance.}
Successful online exploration and conversion into an executable program
do not ensure that its encoded logic covers every situation encountered
during evaluation. A fixed program can respond through its implemented
perception, control, and recovery routines, but situations beyond those
routines may still require an online agent to interpret observations and adapt the next action. The Ethernet insertion results in
Section~\ref{sec:execution_efficiency} illustrate this limitation of the current implementation: successful execution still uses online decisions between programmed motion stages. This suggests a deployment strategy that combines efficient offline execution with online assistance when the program cannot resolve an encountered situation. Determining when to invoke the agent and whether its decisions can subsequently be incorporated into reusable code remains an open question.

\section{Conclusion}
We investigated embodied RSI through real-world robot experiments with
PyRUA-REAL, where coding agents accumulate physical experience as memory
and executable skills.
The real-world experiments demonstrate how physical exploration can produce progressively
improved robot programs that retain acquired capabilities and execute
without repeated agent decisions. This connects recursive self-improvement through
physical interaction to efficient deployment, with progress assessed by
the robust and independence of the resulting programs.
Our findings provide an empirical basis for developing embodied RSI systems that turn physical experience into reliable robot capabilities for robustness deployment.

\clearpage

\bibliographystyle{plainnat}
\setlength{\bibhang}{0pt}
\setlength\bibindent{0pt}
\bibliography{references}

@article{liang2023code,
  title = {{Code as Policies}: Language Model Programs for Embodied Control},
  author = {Liang, Jacky and Huang, Wenlong and Xia, Fei and Xu, Peng and Hausman, Karol and Ichter, Brian and Florence, Pete and Zeng, Andy},
  journal = {arXiv preprint arXiv:2209.07753},
  year = {2022},
  url = {https://arxiv.org/abs/2209.07753}
}

@article{huang2023voxposer,
  title = {{VoxPoser}: Composable 3D Value Maps for Robotic Manipulation with Language Models},
  author = {Huang, Wenlong and Wang, Chen and Zhang, Ruohan and Li, Yunzhu and Wu, Jiajun and Li, Fei-Fei},
  journal = {arXiv preprint arXiv:2307.05973},
  year = {2023},
  url = {https://arxiv.org/abs/2307.05973}
}

@article{huang2024rekep,
  title = {{ReKep}: Spatio-Temporal Reasoning of Relational Keypoint Constraints for Robotic Manipulation},
  author = {Huang, Wenlong and Wang, Chen and Li, Yunzhu and Zhang, Ruohan and Li, Fei-Fei},
  journal = {arXiv preprint arXiv:2409.01652},
  year = {2024},
  url = {https://arxiv.org/abs/2409.01652}
}

@article{fu2026capx,
  title = {{CaP-X}: A Framework for Benchmarking and Improving Coding Agents for Robot Manipulation},
  author = {Fu, Letian and Yu, Justin and El-Refai, Karim and Kou, Ethan and Xue, Haoru and Huang, Huang and Xiao, Wenli and Wang, Guanzhi and Niu, Dantong and Li, Fei-Fei and Shi, Guanya and Wu, Jiajun and Sastry, Shankar and Zhu, Yuke and Goldberg, Ken and Fan, Linxi},
  journal = {arXiv preprint arXiv:2603.22435},
  year = {2026},
  url = {https://arxiv.org/abs/2603.22435}
}

@article{si2026fewer,
  title = {Fewer Tokens, Better Action: {GPT-6 Astra} Robot Agents with 14\% Higher Success Rate but 65\% Fewer Tokens},
  author = {Si, Ruiyang and Bi, Jianxin and Yang, Shunyu and Ni, Rui and Huang, Wenbo and Wang, Qiang and Jiang, Shulong and Zhang, Yintianrun and Ruan, Feixiang and Zhang, Kewei and Wang, Duomin and Li, Xiuyu and Feng, Haiwen and Dong, Zhen and Zhou, Daquan},
  journal = {arXiv preprint arXiv:2610.01939},
  year = {2026},
  url = {https://arxiv.org/abs/2610.01939}
}

@article{jia2026agentaspolicy,
  title = {Agent as Policy for Robotic Manipulation},
  author = {Jia, Mengzhao and Lin, Yang and Zhang, Xixin and Zhang, Zhihan and Liu, Xiaobai and Jiang, Meng},
  journal = {arXiv preprint arXiv:2609.12541},
  year = {2026},
  url = {https://arxiv.org/abs/2609.12541}
}

@article{zhang2026unexpected,
  title = {An Unexpected Robot Policy: Early Evaluations of {GPT-6 Astra} on {RoboDojo} and Beyond},
  author = {Zhang, Wenbo and Wang, Kaixuan and Ouyang, Yutao and Huang, Xiaoyu and Li, Liyang and Su, Kailun and Jin, Weiyang and Chai, Wenhao and Liang, Haotian and Dou, Zhiyang and Chen, Yue and Chen, Tianxing},
  journal = {arXiv preprint arXiv:2609.24170},
  year = {2026},
  url = {https://arxiv.org/abs/2609.24170}
}

@article{sensenova2026embodied,
  title = {Has {GPT-6-Astra} Achieved Embodied Intelligence?},
  author = {{SenseNova} and {Joint Research Team} and Cai, Zhongang and Cao, Ziang and Chen, Tailai and Chua, Han Chong and Dai, Linjun and Deng, Hanming and Deng, Hokin and Diao, Haiwen and Ding, Chenjing and Fan, Weichen and Fu, Xiao and Gong, Zerui and Gu, Chenyang and He, Zexin and Hou, Zhi and Li, Bo and Li, Qinyue and Li, Quanzhou and Li, Xiaowan and Liao, Zhichao and Lin, Dahua and Lin, Zilong and Liu, Bo and Liu, Chunxiao and Liu, Pengfei and Liu, Tianqi and Liu, Ziwei and Loy, Chen Change and Lu, Lewei and Luo, Jiapeng and Ma, Yuexiao and Pan, Liang and Pan, Xingang and Qian, Oscar and Shang, Ziyi and Shao, Yuyao and Su, Dawei and Tao, Dacheng and Tao, Qingyi and Wang, Quan and Wang, Ruisi and Wang, Xiaogang and Wang, Ziyi and Wen, Beichen and Wu, Donghao and Wu, Jiaxu and Wu, Silei and Wu, Xiaoming and Wu, Zhonghua and Xiao, Hongcan and Xu, Weiye and Yamashita, Atsushi and Yang, Lei and Yang, Zhe and Yin, Baiqiao and Yin, Wanqi and Zhao, Hanfei and Zhong, Tianbo and Zhuo, Long and Zuo, Jing},
  journal = {arXiv preprint arXiv:2610.22471},
  year = {2026},
  url = {https://arxiv.org/abs/2610.22471}
}

@article{elmaaroufi2026rho,
  title = {{RHO}: Your Coding Agent is Secretly a Roboticist},
  author = {Elmaaroufi, Karim and Svegliato, Justin and Kalade, Sarunas and Schelle, Graham and Seshia, Sanjit A. and Zaharia, Matei},
  journal = {arXiv preprint arXiv:2606.16458},
  year = {2026},
  url = {https://arxiv.org/abs/2606.16458}
}

@article{liu2026agenticwam,
  title = {Agentic World Action Models},
  author = {Liu, Zhuoyang and Niu, Dantong and Zhang, Botong and Shao, Boning and Wang, Jing and Zheng, Ruijie and Sharma, Yuvan and Lu, Runyu and Liu, Jiaming and Zhang, Renrui and Herzig, Roei and Bai, Yutong and Goldberg, Ken and Xu, Danfei and Zhu, Yuke and Wang, Guanzhi and Fan, Linxi and Darrell, Trevor},
  journal = {arXiv preprint arXiv:2610.30136},
  year = {2026},
  url = {https://arxiv.org/abs/2610.30136}
}

@article{shinn2023reflexion,
  title = {Reflexion: Language Agents with Verbal Reinforcement Learning},
  author = {Shinn, Noah and Cassano, Federico and Berman, Edward and Gopinath, Ashwin and Narasimhan, Karthik and Yao, Shunyu},
  journal = {arXiv preprint arXiv:2303.11366},
  year = {2023},
  url = {https://arxiv.org/abs/2303.11366}
}

@article{zhao2023expel,
  title = {{ExpeL}: {LLM} Agents Are Experiential Learners},
  author = {Zhao, Andrew and Huang, Daniel and Xu, Quentin and Lin, Matthieu and Liu, Yong-Jin and Huang, Gao},
  journal = {arXiv preprint arXiv:2308.10144},
  year = {2023},
  url = {https://arxiv.org/abs/2308.10144}
}

@article{zelikman2023stop,
  title = {Self-Taught Optimizer ({STOP}): Recursively Self-Improving Code Generation},
  author = {Zelikman, Eric and Lorch, Eliana and Mackey, Lester and Kalai, Adam Tauman},
  journal = {arXiv preprint arXiv:2310.02304},
  year = {2023},
  url = {https://arxiv.org/abs/2310.02304}
}

@article{hu2024automated,
  title = {Automated Design of Agentic Systems},
  author = {Hu, Shengran and Lu, Cong and Clune, Jeff},
  journal = {arXiv preprint arXiv:2408.08435},
  year = {2024},
  url = {https://arxiv.org/abs/2408.08435}
}

@article{yin2024godel,
  title = {G{\"o}del Agent: A Self-Referential Agent Framework for Recursive Self-Improvement},
  author = {Yin, Xunjian and Wang, Xinyi and Pan, Liangming and Lin, Li and Wan, Xiaojun and Wang, William Yang},
  journal = {arXiv preprint arXiv:2410.04444},
  year = {2024},
  url = {https://arxiv.org/abs/2410.04444}
}

@article{zhang2025darwin,
  title = {Darwin Godel Machine: Open-Ended Evolution of Self-Improving Agents},
  author = {Zhang, Jenny and Hu, Shengran and Lu, Cong and Lange, Robert and Clune, Jeff},
  journal = {arXiv preprint arXiv:2505.22954},
  year = {2025},
  url = {https://arxiv.org/abs/2505.22954}
}

@article{xiao2026enpire,
  title = {{ENPIRE}: Agentic Robot Policy Self-Improvement in the Real World},
  author = {Xiao, Wenli and Xie, Jia and Zhang, Tonghe and Lin, Haotian and Fu, Letian and Xue, Haoru and Lu, Jalen and Yang, Yi and Dai, Cunxi and Wang, Zi and Wu, Jimmy and Wang, Guanzhi and Sastry, S. Shankar and Goldberg, Ken and Fan, Linxi and Zhu, Yuke and Shi, Guanya},
  journal = {arXiv preprint arXiv:2606.19980},
  year = {2026},
  url = {https://arxiv.org/abs/2606.19980}
}

@article{lu2026aspire,
  title = {{ASPIRE}: Agentic /Skills Discovery for Robotics},
  author = {Lu, Runyu and Wu, Yubo and Kou, Ethan and Lin, Haotian and Fu, Letian and Xiao, Wenli and Mandlekar, Ajay and Xu, Yinzhen and Shi, Guanya and Goldberg, Ken and Chen, Ang and Chowdhury, Mosharaf and Kautz, Jan and Zhu, Yuke and Fan, Linxi and Wang, Guanzhi},
  journal = {arXiv preprint arXiv:2607.00272},
  year = {2026},
  url = {https://arxiv.org/abs/2607.00272}
}

@article{wen2026roborsi,
  title = {{RoboRSI}: Stable, Efficient, and Reusable Robot Self-Evolution in Complex Real-World Environments},
  author = {Wen, Zimo and Chen, Yijin and Cao, Yuxuan and Chen, Wendi and Zou, Yanwen and Yu, Wenye and Kuang, Fuhang and Xue, Han and Lv, Jun and Wen, Chuan and Lu, Cewu},
  journal = {arXiv preprint arXiv:2610.12424},
  year = {2026},
  url = {https://arxiv.org/abs/2610.12424}
}

@article{hu2026embodiedturing,
  title = {Embodied Turing Machines: Stateful Code for Robot Recursive Self-Improvement},
  author = {Hu, Kairui and Hu, Siyuan and Hong, Fangzhou and Chen, Zhaoxi and Liu, Ziwei},
  journal = {arXiv preprint arXiv:2610.12369},
  year = {2026},
  url = {https://arxiv.org/abs/2610.12369}
}

@article{wang2026shaper,
  title = {Self-Evolving Embodied Agents via Skill-Harness Evolution},
  author = {Wang, Peidong and Ma, Zhiming and Chang, Ying and Luo, Xufang and Zhang, Yiqun and Wang, Zihan and Yang, Xiaocui and Feng, Shi and Yang, Yuqing and Li, Dongsheng},
  journal = {arXiv preprint arXiv:2608.11350},
  year = {2026},
  url = {https://arxiv.org/abs/2608.11350}
}

@article{wang2026rpg,
  title = {Reconstruct, Practice, Go Real: Guided Self-Improvement for Embodied Agents},
  author = {Wang, Yen-Jen and Jiang, Haozhe and Deng, Shuying and Xue, Haoru and Ye, Weirui and Duan, Rocky and Haghtalab, Nika and Sastry, S. Shankar and Abbeel, Pieter and Qi, Haozhi},
  journal = {arXiv preprint arXiv:2610.02204},
  year = {2026},
  url = {https://arxiv.org/abs/2610.02204}
}

@article{li2026embodiedevo,
  title = {{EmbodiedEvo}: Self-Evolving Coding Agent for Physical Intelligence},
  author = {Li, Wenyi and Chi, Yizhe and Zheng, Youjie and Liu, Chunyu and Wang, Xiaoqiu and Gao, Mingju and Jiang, Zheng and Hong, Deyao and Xiao, Calvin and Yang, Kaisen and Na, Qinhuai},
  journal = {arXiv preprint arXiv:2610.18867},
  year = {2026},
  url = {https://arxiv.org/abs/2610.18867}
}

@article{song2026embodiedrsi,
  title = {{EmbodiedRSI}: Active Continual Robot Learning Through Hypothesis-Guided Co-Evolution},
  author = {Song, Python and Liang, Zhixuan and Fu, Kelsey and Wang, Mengdi and Yang, Junfeng and Liu, Shilong},
  journal = {arXiv preprint arXiv:2610.10498},
  year = {2026},
  url = {https://arxiv.org/abs/2610.10498}
}

@article{wang2026sear,
  title = {{SEAR}: Self-Evolving Agentic Robotics},
  author = {Wang, Xinyi and Li, Bangzheng and Wang, Hanyu and Zhang, Lin and Cui, Andrea Yaoyun and Tang, Zhenggang and Buch, Shyamal and Xu, Dejia},
  journal = {arXiv preprint arXiv:2610.18792},
  year = {2026},
  url = {https://arxiv.org/abs/2610.18792}
}

@misc{zhai2023sigmoidlosslanguageimage,
      title={Sigmoid Loss for Language Image Pre-Training}, 
      author={Xiaohua Zhai and Basil Mustafa and Alexander Kolesnikov and Lucas Beyer},
      year={2023},
      eprint={2303.15343},
      archivePrefix={arXiv},
      primaryClass={cs.CV},
      url={https://arxiv.org/abs/2303.15343}, 
}

@misc{kirillov2023segment,
      title={Segment Anything}, 
      author={Alexander Kirillov and Eric Mintun and Nikhila Ravi and Hanzi Mao and Chloe Rolland and Laura Gustafson and Tete Xiao and Spencer Whitehead and Alexander C. Berg and Wan-Yen Lo and Piotr Dollár and Ross Girshick},
      year={2023},
      eprint={2304.02643},
      archivePrefix={arXiv},
      primaryClass={cs.CV},
      url={https://arxiv.org/abs/2304.02643}, 
}

@article{jiang2026appl,
  title   = {Agent Priors-guided Policy Learning},
  author  = {Jiang, Puming and Hu, Tianrun and Du, Haozhe and Li, Yibo and
             Xue, Zhiwei and Li, Xinhu and Soh, Harold},
  journal = {arXiv preprint arXiv:2609.35690},
  year    = {2026}
}

\clearpage
\appendix

\section{Execution Feedback: Protocol, Costs, and Diagnostic Traces}
\label{app:feedback_protocol}
\label{app:feedback_analysis}

\paragraph{Information supplied for revision.}
Table~\ref{tab:feedback_conditions} defines the eight conditions evaluated on
all four tasks. The main-text comparison focuses on B, BS, and BVJ to contrast
basic outcomes, numeric state diagnostics, and video with joint trajectories.
BS includes privileged ground-truth object states supplied to the revising
agent. The
remaining conditions and revision histories are reported below.
The persistent optimizer is GPT-6-Astra with medium
reasoning effort. It receives a task description, interface specification,
and a shared calibration observation before generating revision~0.
Each condition uses an isolated workspace and a single retained session
through revision~5; delegation is disabled. Feedback comes only from
development layouts 5 and 6. Test layouts are 32--37, except that \emph{Deposit coin}
layout 35 is replaced by 44 for every condition after a zero-action
initialization failure. Test outcomes are not supplied to the optimizer.

\begin{table}[H]
    \centering
    \small
    \caption{Feedback interfaces. All conditions include B. Video comprises
    three synchronized camera recordings at 25\,Hz, accessed through tools.
    $\dagger$~Privileged diagnostics.}
    \label{tab:feedback_conditions}
    \setlength{\tabcolsep}{5pt}
    \renewcommand{\arraystretch}{1.15}
    \begin{tabularx}{\linewidth}{@{}lX@{}}
        \toprule
        Condition & Information added to the basic report \\
        \midrule
        B & Success, termination reason, action count, simulation time, and episode wall time. \\
        BV & Camera frames at steps 0, 50, 100, 200, 400, and the final available step. \\
        BV\_FULL & Complete camera video, without joint-trajectory CSV. \\
        BVJ & Complete video and synchronized joint-state CSV. \\
        BVJ\_QUERY & BVJ plus an optional frame-query and contact-sheet helper. \\
        BS$^{\dagger}$ & Sampled numeric robot and ground-truth object states. \\
        BESV$^{\dagger}$ & Progress events, state diagnostics, and sampled frames. \\
        BVJ\_ESV$^{\dagger}$ & BVJ plus progress events, state diagnostics, and sampled frames. \\
        \bottomrule
    \end{tabularx}
\end{table}

\paragraph{Execution inputs and scope.}
The programs in this diagnostic study use the same RGB-D, calibration,
and robot-state interface. Object-state feedback, which is part of ground truth, is additionally supplied
to the optimizer in the three marked conditions.

\begin{figure}[H]
    \centering
    \includegraphics[width=\linewidth]{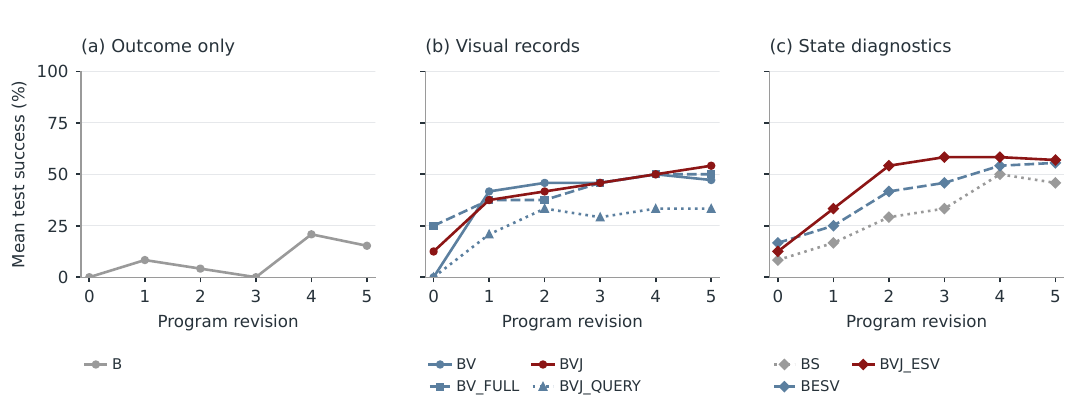}
    \caption{Test success across revisions, averaged equally over four tasks.
    Revisions~0--4 use one execution per test layout; revision~5 uses three.
    Condition codes are defined in Table~\ref{tab:feedback_conditions}.}
    \label{fig:feedback_trajectories}
\end{figure}

\paragraph{Checkpoint and sampling protocol.}
There is one independently initialized exploration session per
task--condition pair. The main comparison contains 32 sessions, each with
one initial generation and five revisions. Every final program is tested
on six layouts with three executions per layout: 18 executions per cell
and 576 in total.
The fixed-revision comparison in Figure~\ref{fig:feedback_results} uses
revision~5, while Figure~\ref{fig:feedback_costs} uses the separate
retrospective selection rule below. Figure~\ref{fig:feedback_trajectories} displays
the full revision trajectories.

\subsection{Observed Development Cost}
\label{app:feedback_cost}
\label{app:feedback_costs}

\begin{figure}[H]
    \centering
    \includegraphics[width=\linewidth]{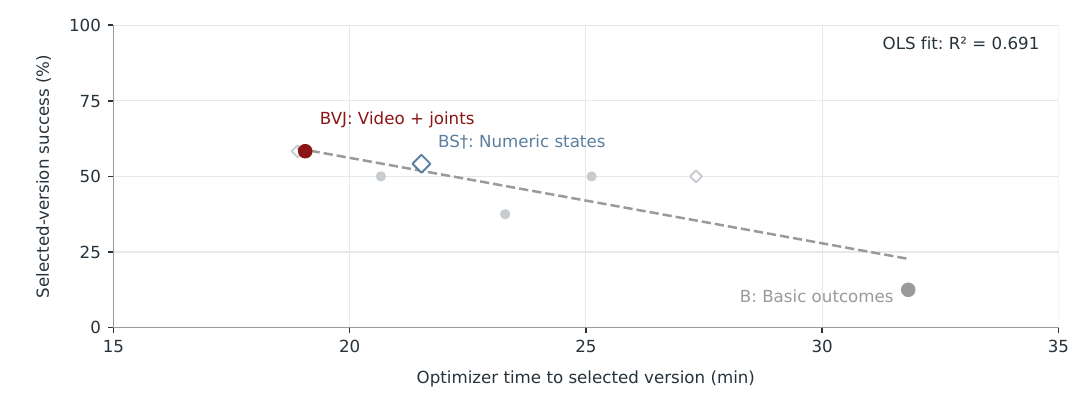}
    \caption{\textbf{Retrospective selected-version success and development cost.}
    Each point averages four tasks under one feedback condition; the dashed
    line is an ordinary least-squares fit across all eight conditions.
    B, BS, and BVJ are labeled; hollow diamonds indicate privileged
    diagnostics. Versions are selected using test outcomes. Time includes
    initial generation through the selected revision, although all
    sessions continue through revision 5.}
    \label{fig:feedback_costs}
    \label{fig:feedback_retrospective_cost}
\end{figure}

\paragraph{Cost through the selected version.}
Figure~\ref{fig:feedback_costs} uses all eight conditions; B, BS, and BVJ
are labeled, while hollow diamonds mark privileged diagnostic conditions.
For each task--condition pair, all revisions~0--5 are scored
on the same six test layouts using the first valid execution per layout,
including at revision~5. If any revision achieves at least 4/6 successes,
we select the highest-scoring revision, breaking ties in favor of the
earliest. Otherwise, we retain revision~5, even if an earlier revision has
a higher score below 4/6. Selection uses test scores, not development
scores; the selected-version scores are therefore retrospective summaries.

Costs sum recorded optimizer turns, including retries, from initial
generation through the selected revision. Each plotted coordinate is the
arithmetic mean over four tasks. The ordinary least-squares fit across
the eight condition means has $R^2=0.691$.

All sessions continue through revision~5. Selected-version costs measure
work accumulated through the selected revision; Table~\ref{tab:feedback_cost}
reports these costs alongside the full exploration costs.

Across the eight condition means, selected-version success is associated
with lower cumulative optimizer time ($R^2=0.691$). With video and joint
trajectories, selected-version success is 58.3\%, compared with 12.5\%
for basic outcomes; cumulative optimizer time is 19.1 versus 31.8 minutes
and tokens are 2.43 versus 3.54 million. Because selection uses test
outcomes, these retrospective summaries do not establish that an agent
could identify the selected version or stop at it online.

\paragraph{Full exploration cost and measurement scope.}
Full exploration costs include initial generation and all five revisions.
Optimizer time is the sum of agent-turn durations,
including tools invoked within those turns. Separate simulator wall time
covers development evaluations, excluding queueing. Tokens are total input
plus output tokens accumulated across turns, including cached input. Test
evaluation costs are excluded.

\begin{table}[H]
    \centering
    \footnotesize
    \caption{Selected-version and full exploration costs, averaged equally
    over four tasks. Selected revisions are ordered as \emph{Stack blocks},
    \emph{Push T}, \emph{Deposit coin}, and \emph{Imitate sorting sequence}. Selected success uses one execution
    per test layout; final success uses three. Times are in minutes and tokens
    in millions. $\dagger$~Privileged diagnostics.}
    \label{tab:feedback_cost}
    \setlength{\tabcolsep}{3pt}
    \renewcommand{\arraystretch}{1.15}
    \begin{tabular*}{\linewidth}{@{\extracolsep{\fill}}llrrrrrrr@{}}
        \toprule
        & & \multicolumn{3}{c}{Through selected revision}
            & \multicolumn{4}{c}{Full exploration (r0--r5)} \\
        \cmidrule(lr){3-5} \cmidrule(l){6-9}
        Condition & Selected revisions & \shortstack{Success\\(\%)}
            & \shortstack{Optimizer\\time} & Tokens
            & \shortstack{Success\\(\%)} & \shortstack{Optimizer\\time}
            & Tokens & \shortstack{Simulator\\time} \\
        \midrule
        B & 5 / 5 / 5 / 5 & 12.50 & 31.82 & 3.536 & 15.28 & 31.82 & 3.536 & 29.63 \\
        BV & 1 / 4 / 5 / 5 & 50.00 & 25.12 & 2.747 & 47.22 & 29.41 & 3.417 & 25.14 \\
        BV\_FULL & 3 / 1 / 5 / 5 & 50.00 & 20.66 & 2.607 & 50.00 & 24.40 & 3.606 & 28.86 \\
        BVJ & 2 / 1 / 5 / 5 & 58.33 & 19.06 & 2.433 & 54.17 & 25.40 & 3.647 & 27.28 \\
        BVJ\_QUERY & 5 / 2 / 5 / 5 & 37.50 & 23.29 & 3.836 & 33.33 & 25.42 & 4.356 & 32.43 \\
        BS$^{\dagger}$ & 1 / 4 / 5 / 5 & 54.17 & 21.52 & 2.203 & 45.83 & 24.98 & 2.714 & 28.35 \\
        BESV$^{\dagger}$ & 4 / 3 / 5 / 5 & 50.00 & 27.33 & 3.607 & 55.56 & 29.20 & 4.010 & 27.87 \\
        BVJ\_ESV$^{\dagger}$ & 2 / 3 / 5 / 5 & 58.33 & 18.90 & 3.255 & 56.94 & 21.34 & 3.893 & 30.30 \\
        \bottomrule
    \end{tabular*}
\end{table}

\subsection{Evidence Actually Inspected in Physical Tasks}
\label{app:feedback_real}

A physical failure report does not by itself identify what should change.
The traces below show which observations the agent inspected, the revision
it made, and the check that followed. We distinguish direct image inspection
from diagnosis recorded in experiment logs.

\paragraph{Opening a failed placement image reveals a detector error.}

\begin{figure}
\centering
\includegraphics[width=0.9\linewidth]{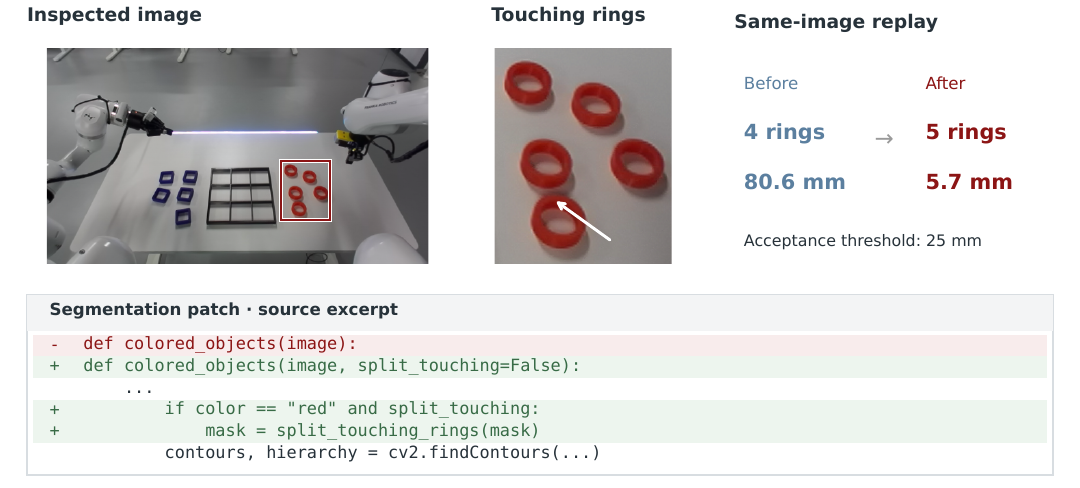}
\caption{\textbf{From an inspected camera image to a verified code repair.}
The agent inspected the original photo; the crop and arrow are added for
readability. The frozen-source diff omits unchanged preprocessing and wraps
the added conditional; red and green mark removed and added code.
The replay changes verification, not robot motion.}
\label{fig:physical_inspection_rings}
\end{figure}

During tic-tac-toe setup, a red-ring transfer is rejected because its reported
placement error is 80.6\,mm, above the 25\,mm threshold. The engineer reads \texttt{execution.json} and calls \texttt{view\_image} on
\texttt{head-003-placed.png} in the same tool invocation. The image contains
five red rings, but the execution record contains only four detections:
two touching rings share one contour. The mismatch directs the revision
to segmentation. Figure~\ref{fig:physical_inspection_rings} shows the actual
image and an excerpt of the diff between the frozen pre- and post-fix sources.
The new helper splits an outer contour only when it contains exactly two
holes larger than 60 pixels in area, using the bisector of their centers.

The agent reprocesses the same image and obtains five detections; the nearest
ring is now 5.7\,mm from the target under the unchanged threshold. It then
runs 50 regression checks, including the failure image and obstruction
rejection, and checks 178 fresh static perception frames. After restarting
the physical task, it reads the new execution events and confirms the initial
stock transfer and the first blue and red placements. This establishes a
verifier repair; the same-image comparison is not a new physical placement,
and the 178 frames are not 178 manipulation trials.

\begin{figure}
\centering
\includegraphics[width=0.8\linewidth]{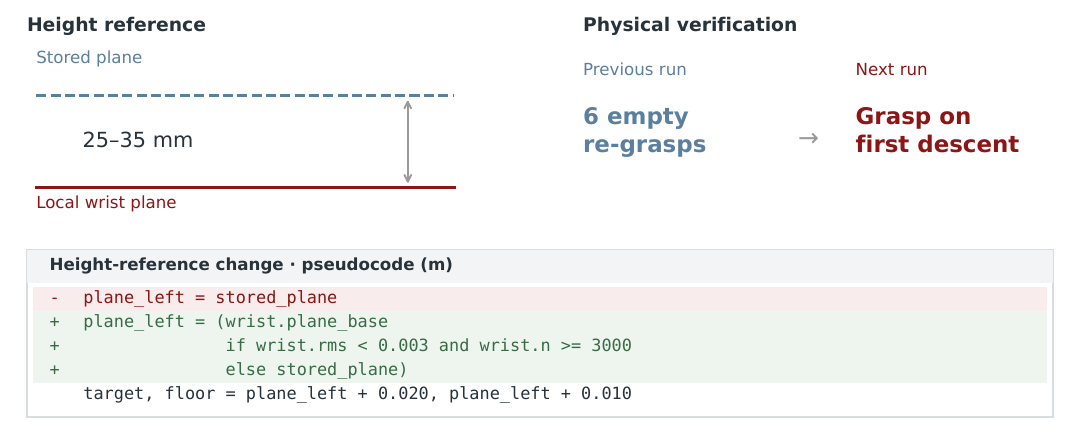}
\caption{\textbf{From empty grasps to a revised physical descent.}
The plane mismatch explains the empty grasps; the next recorded unfold
attempt captures the towel on its first descent. The drawing is not to scale.
The code comparison is pseudocode reconstructed from the iteration record,
not a recovered source diff.}
\label{fig:physical_inspection_towel}
\end{figure}

\paragraph{Depth and gripper logs reveal why towel grasps close in the air.}
A towel-unfolding trace provides a complementary change to physical execution
(Figure~\ref{fig:physical_inspection_towel}). In \texttt{cycle\_142100}, cycle~4,
the left gripper exhausts six re-grasp attempts and aborts. The agent's
iteration record compares the stored table plane with the local wrist-depth
fit and reconstructs pad heights from the logged arm states. Across the
failed cycles, the stored plane is 25--35\,mm higher than the local fit,
so its descent floor can leave the pads above the cloth.

The recorded revision uses the left wrist's local table plane when its fit
has RMS error below 3\,mm and at least 3,000 points. It retains the force probe
(4\,N rise trigger, 10\,N hard limit) and expresses the descent target and
re-grasp floor relative to that plane.
In the next run, \texttt{cycle\_143600}, cycle~1, the left gripper captures the
towel on the first descent, with a 1.4\,mm RMS fit over 6,496 points.
The log reports both grippers closed and an
unfolded extent of $0.30\times0.42$\,m. This is one recorded unfold-stage
recovery, not a repeated full-task success rate. The following cycle stops
on a connection error. Here the evidence is the raw execution logs and the
agent's contemporaneous diagnosis; no corresponding image-open call is
claimed.

\FloatBarrier

\clearpage
\section{Generalization under Explicit Task Requirements: Per-Task Results}
\label{app:generalization}

Table~\ref{tab:generalization_per_task} reports the success rates underlying
the category means in Figure~\ref{fig:generalization_survey}.
Both policies are evaluated on identical hidden layouts; layouts used to
select the baseline policy are excluded. Programs receive RGB-D observations,
including simulator-rendered depth.

\begin{table}[H]
\centering
\small
\caption{Per-task hidden-layout success rates before and after survey-assisted improvement. $\Delta$ is the absolute change in percentage points. Tested on 32 tasks.}
\label{tab:generalization_per_task}
\setlength{\tabcolsep}{7pt}
\renewcommand{\arraystretch}{1.15}
\begin{tabularx}{\linewidth}{@{}Xrrr@{}}
\toprule
Task & Before (\%) & After (\%) & $\Delta$ (pp) \\
\midrule
\multicolumn{4}{@{}l}{\textit{Generalization}} \\
Arrange largest number & 52.6 & 94.7 & $+42.1$ \\
Fold clothes & 40.9 & 59.1 & $+18.2$ \\
Hang mugs & 10.5 & 26.3 & $+15.8$ \\
Make toast & 28.9 & 28.9 & $0.0$ \\
Pack objects into box & 2.3 & 6.8 & $+4.5$ \\
Pour liquid into cup & 59.1 & 100.0 & $+40.9$ \\
Push T & 52.3 & 70.5 & $+18.2$ \\
Sort nesting dolls by size & 15.8 & 78.9 & $+63.2$ \\
Stack bowls & 0.0 & 86.4 & $+86.4$ \\
Store laptop and headphones & 0.0 & 0.0 & $0.0$ \\
\addlinespace[3pt]
\multicolumn{4}{@{}l}{\textit{Precision}} \\
Deposit coin & 28.6 & 64.3 & $+35.7$ \\
Fasten screws & 9.1 & 81.8 & $+72.7$ \\
Insert key & 0.0 & 18.2 & $+18.2$ \\
Insert tubes & 79.5 & 88.6 & $+9.1$ \\
Plug in charger & 11.4 & 22.7 & $+11.4$ \\
Pour balls into vase & 54.5 & 77.3 & $+22.7$ \\
\addlinespace[3pt]
\multicolumn{4}{@{}l}{\textit{Long-Horizon}} \\
Classify objects & 48.8 & 79.1 & $+30.2$ \\
Fill egg holder & 0.0 & 35.7 & $+35.7$ \\
Fill pen holder & 4.5 & 13.6 & $+9.1$ \\
Organize table & 6.8 & 15.9 & $+9.1$ \\
Play tic-tac-toe & 82.2 & 100.0 & $+17.8$ \\
Put bottles into dustbin & 4.5 & 77.3 & $+72.7$ \\
\addlinespace[3pt]
\multicolumn{4}{@{}l}{\textit{Memory}} \\
Imitate sorting sequence & 28.9 & 31.1 & $+2.2$ \\
Match and pick from conveyor & 31.8 & 72.7 & $+40.9$ \\
Swap T & 86.4 & 97.7 & $+11.4$ \\
Swap blocks & 68.2 & 59.1 & $-9.1$ \\
\addlinespace[3pt]
\multicolumn{4}{@{}l}{\textit{Open}} \\
Classify objects by language & 25.0 & 50.0 & $+25.0$ \\
General pickup & 0.0 & 74.4 & $+74.4$ \\
Pick from conveyor by image & 18.2 & 18.2 & $0.0$ \\
Pour by language & 4.5 & 70.5 & $+65.9$ \\
Solve equation & 38.6 & 75.0 & $+36.4$ \\
Store tools in toolbox & 0.0 & 18.2 & $+18.2$ \\
\bottomrule
\end{tabularx}
\end{table}

\FloatBarrier
\clearpage

\section{Physical Task Configurations and Completion Criteria}
\label{app:task_settings}

\paragraph{Tic-tac-toe.}
The two arms alternate legal moves on a $3\times3$ board. The left arm places blue square-frame pieces and the right arm places red rings. A game terminates with a win or draw. The complete-cycle measurements also include clearing the board and returning the pieces to their respective sides.

\paragraph{Five-block stacking.}
Five blocks initially lie on the table. The robot stacks the blocks to form a freestanding tower. Complete-cycle measurements include dismantling the tower in reverse order and returning the four upper blocks to their table positions.

\paragraph{Tower of Hanoi.}
The robot sorts eight disks by size and places them onto the middle peg, from largest at the bottom to smallest at the top. This task evaluates ordered disk manipulation and repeated hole--peg alignment, rather than solving the conventional three-peg transfer puzzle. Complete-cycle measurements also include removing the disks and returning them to the table.

\paragraph{Ethernet insertion.}
The robot grasps a yellow-booted RJ45 connector from the table and inserts it into the Ethernet port of a fixed computer chassis. Completion requires the connector to remain seated after the gripper releases it and retreats. Electrical connectivity is not evaluated.

\paragraph{Towel folding.}
The towel initially lies flat on the table. The two arms grasp it and fold it over onto itself, requiring coordinated manipulation of a deformable object. Evaluation covers the folding operation only.

\paragraph{Push T.}
The robot pushes a T-shaped object across the table to align its position and orientation with a target pose. Manipulation is performed through pushing, without grasping or lifting the object.

\paragraph{Block sorting.}
The robot identifies orange and dark-blue blocks and places them into their corresponding color bins. Each trial covers sorting the specified blocks and combines visual discrimination with repeated pick-and-place operations.

\paragraph{Block storage.}
The robot collects scattered blocks into a bowl. Success requires storing all blocks. Missed grasps followed by successful recovery remain part of the same trial.

\clearpage
\section{Full RoboDojo Results}
\label{app:simulation_results}

The official PyRUA-REAL values are documented in its \href{https://github.com/DAGroup-PKU/PyRUA-Lean/blob/main/integrations/xpolicylab/evaluation/robodojo_summary_rsi.md}{evaluation summary}. Its separately reported self-run is not used here. Evaluation seeds for the other baseline rows are not specified in the leaderboard snapshot.

\begin{table}[htbp]
\centering
\footnotesize
\setlength{\tabcolsep}{3pt}
\caption{Full RoboDojo comparison from the leaderboard snapshot's October 24,
2026 leaderboard snapshot. Capability columns report success rates
(\%); the final two columns weight the five capabilities equally.
Best values in each column are bold. These are reported results,
not a matched-seed comparison: our evaluation uses seed 0, whereas
the official PyRUA-REAL evaluation uses seed 3.}
\label{tab:full_robodojo_results}
\begin{tabular}{lrrrrrrr}
\hline
Method & Gen. & Prec. & Long. & Memory & Open & Avg. SR & Avg. score \\
\hline
\multicolumn{8}{l}{\textit{Agent}} \\
PyRUA-Lean & 43.17 & 38.25 & 43.75 & 68.33 & 58.25 & 50.35 & 55.26 \\
HexaShell-Preview & 34.50 & 32.00 & 31.50 & 81.00 & 55.25 & 46.85 & 53.18 \\
EmbodiedEvo & 30.83 & 16.25 & 21.75 & 76.67 & 49.25 & 38.95 & 45.08 \\
GPT-6-Astra & 30.50 & 4.00 & 8.25 & 38.67 & 31.00 & 22.48 & 28.97 \\
DeepSeek-Flash & 1.67 & 0.00 & 0.00 & 1.67 & 6.25 & 1.92 & 2.99 \\
GPT-5.5 & 0.00 & 0.00 & 0.00 & 1.67 & 2.75 & 0.88 & 1.13 \\
\hline
\multicolumn{8}{l}{\textit{Agent hybrids}} \\
WorldScape-R2 SI (+WAM) & 35.83 & 15.50 & 34.25 & 63.67 & 47.50 & 39.35 & 46.58 \\
PhysicalRSI (+VLA) & 15.62 & 32.50 & 37.33 & 46.56 & 24.92 & 31.39 & 36.27 \\
\hline
\multicolumn{8}{l}{\textit{Ours}} \\
PyRUA-REAL (ours) & \textbf{72.33} & \textbf{91.50} & \textbf{80.00} & \textbf{95.55} & \textbf{88.24} & \textbf{84.42} & \textbf{88.53} \\
\hline
\end{tabular}
\end{table}

\end{document}